\documentclass[12pt]{article}
\usepackage{amsmath, amsfonts,dsfont,mathrsfs,amssymb}
\usepackage{graphicx}
\usepackage{enumerate}
\usepackage[semicolon]{natbib}
\usepackage{url} 
\usepackage{hyperref}
\usepackage{xcolor}
\usepackage{pifont}
\usepackage{bm}
\usepackage{appendix}
\usepackage{multirow}
\usepackage{comment}
\usepackage{rotating}
\usepackage{float}
\usepackage{multibib}
\usepackage{lscape}
\usepackage{makecell}
\usepackage{fancyhdr}
\usepackage{indentfirst}

\newcites{sm}{Supplementary References}

\renewcommand{\and}{\textcomma~}
\newcommand{\blind}{0}

\newcommand{\red}{\color{red}}

\newcommand{\gray}{\color{gray}}

\newcommand{\bA}{{\bf A}}
\newcommand{\bB}{{\bf B}}

\newcommand{\bP}{{\bf P}}
\newcommand{\bT}{{\bf T}}
\newcommand{\bX}{{\bf X}}
\newcommand{\bY}{{\bf Y}}

\newcommand{\bV}{{\bf V}}

\newcommand{\bs}{{\bf s}}

\newcommand{\balpha}{{\bm \alpha}}
\newcommand{\bbeta}{{\bm \beta}}
\newcommand{\blambda}{{\bm \lambda}}

\newcommand{\bphi}{{\bm \phi}}
\newcommand{\bpsi}{{\bm \psi}}

\newcommand{\btheta}{\boldsymbol{\theta}}

\newcommand{\bbP}{\mathbb{P}}

\newcommand{\bbI}{\mathbb{I}}
\newcommand{\bbN}{\mathbb{N}}
\newcommand{\bbE}{\mathbb{E}}

\newcommand{\bbU}{\mathbb{U}}
\newcommand{\bbV}{\mathbb{V}}

\newcommand{\bbS}{\mathbb{S}}

\newcommand{\cF}{{\cal F}}
\newcommand{\cI}{{\cal I}}

\newcommand{\cA}{\mathcal{A}}
\newcommand{\cB}{\mathcal{B}}
\newcommand{\cC}{\mathcal{C}}
\newcommand{\cD}{\mathcal{D}}
\newcommand{\cL}{\mathcal{L}}
\newcommand{\cM}{\mathcal{M}}

\newcommand{\cP}{\mathcal{P}}
\newcommand{\cS}{\mathcal{S}}
\newcommand{\cT}{\mathcal{T}}
\newcommand{\cV}{\mathcal{V}}

\makeatletter

\newcommand{\Rmnum}[1]{\expandafter\@slowromancap\romannumeral #1@}
\makeatother

\begin{document}

\def\spacingset#1{\renewcommand{\baselinestretch}%
{#1}\small\normalsize} \spacingset{1}


\if0\blind
{
    \title{\bf Efficient Surgical Tool Recognition via HMM-Stabilized Deep Learning}
    \author{Haifeng Wang\textsuperscript{\rm 1,$\ast$},
    Hao Xu\textsuperscript{\rm 2,$\ast$},
    Jun Wang\textsuperscript{\rm 2},
    Jian Zhou\textsuperscript{\rm 2,$\dagger$},
    Ke Deng\textsuperscript{\rm 1,$\dagger$}\\
    \\
    \textsuperscript{\rm 1}Center for Statistical Science \& Department of Industrial Engineering,\\ Tsinghua University, Beijing, China \\
    \textsuperscript{\rm 2}Department of Thoracic Surgery, Peking University People's Hospital,\\ Beijing, China\\
}
    \date{}
    \maketitle
} \fi

\if1\blind
{
  \bigskip
  \bigskip
  \bigskip
  \begin{center}
    {\LARGE\bf Efficient Surgical Tool Recognition via HMM-Stabilized Deep Learning}
\end{center}
  \medskip
} \fi

\bigskip
\begin{abstract}
Recognizing various surgical tools, actions and phases from surgery videos is an important problem in computer vision with exciting clinical applications.
Existing deep-learning-based methods for this problem either process each surgical video as a series of independent images without considering their dependence, or rely on complicated deep learning models to count for dependence of video frames.
In this study, we revealed from exploratory data analysis that surgical videos enjoy relatively simple semantic structure, where the presence of surgical phases and tools can be well modeled by a compact hidden Markov model (HMM).
Based on this observation, we propose an HMM-stabilized deep learning method for tool presence detection.
A wide range of experiments confirm that the proposed approaches achieve better performance with lower training and running costs, and support more flexible ways to construct and utilize training data in scenarios where not all surgery videos of interest are extensively labelled.
These results suggest that popular deep learning approaches with over-complicated model structures may suffer from inefficient utilization of data, and integrating ingredients of deep learning and statistical learning wisely may lead to more powerful algorithms that enjoy competitive performance, transparent interpretation and convenient model training simultaneously.
\end{abstract}

\noindent%
\vfill

\newpage
\spacingset{1.9} 

\let\thefootnote\relax\footnotetext{$\ast$ These authors contributed equally.}
\let\thefootnote\relax\footnotetext{$\dagger$ Ke Deng and Jian Zhou are the corresponding authors.}

\section{Introduction}

With the popularization of visual-guided minimally invasive surgeries,
more and more high-quality surgical videos have been naturally generated in operating rooms.
Recording complete operating procedures of various surgeries, these surgical videos contain rich information about surgical progress, providing us a great opportunity to learn and analyze surgical progresses from various aspects with many exciting applications, including surgical quality management \citep{application1}, guidance for junior surgeons \citep{application4}, reference for future surgical phase recognition \citep{endonet,mtrcnet}, or, even more steps further, automated surgical documentation for legal purposes \citep{b5}.
Because it is extremely inefficient for human to process videos by eyeballing, there is a great appeal to develop intelligent algorithms for this purpose.
A primary task in surgical video analysis is \emph{surgical tool recognition} (STR), also known as \emph{tool presence detection} (TPD) in literature, which aims to highlight the presence of a particular collection of surgical tools in each frame of target videos \citep{endonet}.
A closely related but more challenging task is \emph{surgical tool localization}, which goes beyond the TPD task by trying to highlight the concrete location of every appeared tool with a bounding box  \citep{frcnntool}.
Moreover, because a typical surgery often contains multiple phases composed of various surgical actions, researchers also concern the task of \emph{surgical phase/action recognition}, which aims to recognize the surgical phase/action that each frame belongs to. 
In this paper, we mainly focus on the TPD task, but would also consider the joint analysis of these closely related tasks for improved performance.

As a special object recognition problem, surgical tool recognition can be resolved under the general framework of object recognition via transfer deep learning.
In recent years, many methods for surgical tool recognition based on transfer deep learning have been developed in the literature \citep{endonet,fcnnet,frcnntool,mtrcnet,convlstm}.
These methods utilize a pre-trained CNN (often referred to as the ``backbone CNN") to extract frame-level characteristics of the target videos into embedding vectors, and formulate the surgical tool recognition problem as an image classification problem based on the extracted frame-level embedding vectors.
More recently, \cite{Rendezvous, SAIS} consider the more challenging surgical action recognition problem, which involves surgical tool recognition as an intermediate step.
These methods rely on the \emph{attention mechanism} pioneered by \cite{attention} to recognize various surgical actions from surgical videos by learning specific interaction between different surgical tools that appear in the same frame of the target video.

While achieving the state-of-the-art performance, these deep-learning-based methods are often criticized for their black-box nature, lack of interpretability, and unstable performance \citep{dp_interpretability,dp_performance}.
On the other hand, the great success of large deep learning models in this and other fields has encouraged researchers to pursuit larger models for better performance.
However, the fundamental principle of statistical modelling tells us that over-flexible large models without a clear focus on the unique features of the problem of interest would often lead to inefficient utilization of data and sub-optimal results.
In this study, we found via exploratory data analysis that the semantic structure of a surgical video is transparent and highly predictable, with little interactions among frames temporally far away from each other.
Such a fact suggests that a compact \emph{Hidden Markov Model} (HMM) \citep{HMM} is good enough to count for the temporal dependence of surgical video frames, and the sophisticated deep learning architecture utilized by existing methods is definitely over-flexible.
Motivated by this discovery, we propose in this paper an HMM-stabilized deep learning approach for the TPD problem, which integrates statistical learning and deep learning wisely according to the unique feature of surgical videos.
A wide range of experiments confirmed that the proposed HMM-stabilized deep learning achieves competitive recognition accuracy with respect to existing approaches purely based on deep learning, and enjoys many extra advantages, including transparent interpretation, low operating cost, and flexible utilization of training data.

The rest part of this paper is organized as follows.
Section~\ref{sec:Related Works} provides a comprehensive review on existing methods for surgical tool recognition in the literature with discussions on their limitations. 
In Section~\ref{sec:DataSets}, we introduce three surgical video datasets and demonstrate their unique features via exploratory data analysis. 
To overcome the limitations of existing methods, HMM-stabilized deep learning is proposed in Section~\ref{sec:HMM} to integrate the advantages of statistical learning and deep learning.
Performance of the proposed HMM-stabilized deep learning is evaluated in Section~\ref{sec:Experiments} via a series of carefully designed experiments.
Finally, we conclude our study with a discussion in Section~\ref{sec:conclusion}.

\section{Related Works}\label{sec:Related Works}
Pioneered by \cite{endonet}, deep-learning-based approaches for surgical tool recognition have become primary tools for surgical video analysis.
Existing methods in literature can be roughly divided into two categories: the frame-independent methods \citep{endonet,fcnnet,frcnntool} which process frames of a surgical video as independent images via various CNNs for object recognition, and the frame-dependent methods \citep{mtrcnet,convlstm,LapFormer,LAST} which treat a surgical video as a sequence of temporally dependent frames with their temporal correlation delicately considered via additional deep-learning architecture for sequence data analysis.
The frame-dependent recognizers typically performance better than the frame-independent ones at the price of more complex model architecture and higher model training cost. 
Table~\ref{tab:MethodSummary} summarizes the key features of these existing methods, with detailed architecture of some of them illustrated in Section S1.4 of Supplementary Materials.

\subsection{Frame-Independent Methods}
{\bf ToolNet \& PhaseNet.} 
Proposed by \cite{endonet}, ToolNet is one of the earliest deep-learning-based approaches for surgical tool recognition in literature.
Digesting frames of the target surgical videos as independent images, ToolNet utilizes the AlexNet \citep{alexnet} as the backbone CNN to learn an embedding vector for every input frame, and relies on an additional \emph{fully connected layer} (FCL) to predict the presence of each surgical tool of interest in every frame.
Parameters of the AlexNet were pre-trained based on the ImageNet dataset \cite{imagenet}.
Parameters of the FCL were trained based on surgical videos with tool presence labels, as in the Cholec80 dataset, 
under the guidance of a multi-label logistic loss.
The counterpart of ToolNet in surgical phase recognition is PhaseNet \citep{endonet}, which shares a similar architecture as ToolNet with the prediction target changed from tool presence to phase.

{\bf EndoNet.} Also proposed by \cite{endonet}, 
EndoNet extends ToolNet by considering the surgical tool recognition task and the surgical phase recognition task simultaneously.
Enhancing ToolNet with an additional fully connected layer for surgical phase prediction based on the same embedding vector obtained from the AlexNet for each frame, EndoNet outputs probabilistic predictions for surgical tool presence and surgical phase at the same time.
Parameters of EndoNet are trained in a similar way as in ToolNet, with an extra softmax cross-entropy loss for surgical phase prediction.
Because EndoNet makes use of additional information on surgical phases for model training, it performs better than ToolNet and PhaseNet. 

{\bf FR-CNN \& FCN.} Alternatively, surgical tool recognition can be achieved as a byproduct of surgical tool localization, a more challenging task aiming at highlighting the concrete locations of surgical tools in each frame of the target surgery videos with bounding boxes, instead of just predicting their presence.
In literature, object detection tasks such as surgical tool localization can be effectively achieved by R-CNN (\emph{regions with CNN features}) \citep{rcnn}, which generates region proposals of the input images wisely via a selective search procedure 
and evaluates the proposed regions effectively to detect objects of interest.
\cite{frcnntool} trained Faster R-CNN \citep{frcnn}, a classic R-CNN approach for object detection, resulting in an alternative frame-independent surgical tool recognizer that can detect not only presences but also concrete locations of various surgical tools.
Hereinafter, we refer to the above Faster R-CNN method for tool presence detection as FR-CNN.
FR-CNN outperforms ToolNet and EndoNet with a significant margin, at the price of more data annotation efforts (to highlight the bounding boxes) and higher model training cost.
The similar idea was implemented in a slightly different way by \cite{fcnnet}, leading to a surgical tool recognizer with similar properties known as FCN. 

\subsection{Frame-Dependent Methods}
{\bf ToolNet$_L$, EndoNet$_L$ \& FCN$_L$.} A critical limitation of the above frame-independent recognizers is that they all process frames of a surgical video one by one separately without considering their temporal dependence. 
\cite{mtrcnet} filled in this gap by upgrading the surgical phase detection component of EndoNet with an additional LSTM layer to incorporate the temporal dependence among frames in the same video, resulting in a LSTM-enhanced EndoNet named MTRC-Net. 
Similarly, \cite{convlstm} upgraded FCN with an additional LSTM layer to connect the individual localization maps output by the backbone CNN for adjacent frames, resulting in a LSTM-enhanced FCN named ConvLSTM.
In this paper, we refer to MTRC-Net and ConvLSTM as EndoNet$_L$ and FCN$_L$ respectively to highlight their nature as the LSTM-enhanced version of EndoNet and FCN.
In principle, ToolNet and FR-CNN can also be enhanced in the similar way as in EndoNet$_L$ and FCN$_L$ for improved performance, leading to ToolNet$_L$ and FR-CNN$_L$, the LSTM-enhanced version of ToolNet and FR-CNN. 
To the best of our knowledge, however, no research papers had implemented such ideas, leaving ToolNet$_L$ and FR-CNN$_L$ absent in literature.
In this study, we implemented ToolNet$_L$ by ourselves as a complement of EndoNet$_L$ and FCN$_L$.
FR-CNN$_L$ is still absent due to non-trivial issues in implementation.

{\bf ToolNet$_A$, EndoNet$_A$ \& SwinNet$_A$.}
Considering that network architectures with attention mechanism, e.g., transformer \citep{attention} and swin-transformer \citep{swintransformer}, are more powerful to handle complex dependence structure in time series and images, it is a natural idea to further improve the above LSTM-enhanced methods by updating the traditional LSTM layer and/or backbone CNN with attention-based architectures.
\cite{LapFormer} updated ToolNet$_L$ to ToolNet$_A$ (referred to as LapFormer in the original paper) by modelling the temporal dependence of surgical video frames via attention mechanism, and replacing the old-fashion AlexNet backbone with the more capable ResNet50 \citep{resnet}.
\cite{LAST} updated EndoNet$_L$ to EndoNet$_A$ by modelling the temporal dependence of surgical video frames via attention mechanism.
Further improving EndoNet$_A$ by replacing its ResNet50 backbone with a more powerful Swin-B backbone equipped with swin-transformer, \cite{LAST} came up with LAST, the state-of-the-art method in the field.
In this paper, we refer to LAST as SwinNet$_A$ to highlight its nature as an attention-based method with Swin-B as backbone. 
If we downgrade the temporal dependence model with attention mechanism in SwinNet$_A$ to LSTM, or remove it completely, we would obtained two weaker versions of SwinNet$_A$, namely SwinNet$_L$ and SwinNet, according to our rule to name methods.
In this study, we implemented SwinNet and SwinNet$_L$, which have not been formally studies in literature yet, by ourselves.
In principle, the similar idea can also be used to update FR-CNN$_L$ and FCN$_L$ to FR-CNN$_A$ and FCN$_A$.
Unfortunately, however, these three methods are still absent in literature possibly due to challenges in implementation.

\begin{table*}[h]\footnotesize
\renewcommand{\arraystretch}{0.7}
\centering
\resizebox{\columnwidth}{!}{
\begin{tabular}{ccc cc c c}
\hline										
	&&	  \multirow{3}{*}{$\begin{array}{c} \vspace{-0.3cm}\text{Model for}\\ \vspace{0.6cm}\text{frame dependence}\end{array}$}    	&	\multicolumn{2}{c}{Tool}	&&	Phase\\ \cline{4-5}\cline{7-7}
    Method	&   Backbone CNN&&	Presence	&	Localization	&&	Classification\\ 
\hline											
    ToolNet \citep{endonet}	&	AlexNet	&  - &	$\checkmark$	&	$\times$	&&	$\times$	\\																
    EndoNet \citep{endonet}	&	AlexNet	&  - &	$\checkmark$	&	$\times$	&&	$\checkmark$	\\																
    FR-CNN \citep{frcnntool}	&	VGG16	&  - &	$\checkmark$	&	$\checkmark$	&&	$\times$	\\ 																
    FCN \citep{fcnnet}	&	ResNet18	&  - &	$\checkmark$	&	$\checkmark$	&&	$\times$	\\																
    SwinNet (this paper)	&	Swin-B	&  - &	$\checkmark$	&	$\times$	&&	$\checkmark$	\\ \hline																
    ToolNet$_L$ (this paper)	&	AlexNet	&  LSTM &	$\checkmark$	&	$\times$	&&	$\times$	\\																
    EndoNet$_L$\citep{mtrcnet}	&	AlexNet	&  LSTM &	$\checkmark$	&	$\times$	&&	$\checkmark$	\\																
    \gray FR-CNN$_L$ (absent)	&	\gray VGG16	&  \gray LSTM &	\gray $\checkmark$	&	\gray $\checkmark$	&&	\gray $\times$	\\ 																
    FCN$_L$\citep{convlstm} 	&	ResNet18	&  LSTM &	$\checkmark$	&	$\checkmark$	&&	$\times$	\\ 																
    SwinNet$_L$ (this paper)	&	Swin-B	&  LSTM &	$\checkmark$	&	$\times$	&&	$\checkmark$	\\\hline																
    ToolNet$_A$ \citep{LapFormer}	&	ResNet50 &  Attention &	$\checkmark$	&	$\times$	&&	$\times$	\\																
    EndoNet$_A$ \citep{LAST}	&	ResNet50 &  Attention &	$\checkmark$	&	$\times$	&&	$\checkmark$	\\																
    \gray FR-CNN$_A$ (absent)	&	\gray VGG16	&  \gray Attention &	\gray $\checkmark$	&	\gray $\checkmark$	&&	\gray $\times$	\\																
    \gray FCN$_A$(absent) 	&	\gray ResNet18	&  \gray Attention &	\gray $\checkmark$	&	\gray $\checkmark$	&&	\gray $\times$	\\ 																
    SwinNet$_A$ \citep{LAST}	&	Swin-B	&  Attention &	$\checkmark$	&	$\times$	&&	$\checkmark$	\\ \hline
\end{tabular}
}
\caption{Summary of existing deep-learning-based methods for surgical video analysis.}
\label{tab:MethodSummary}
\end{table*}

\section{Data}\label{sec:DataSets}
In this study, we consider 3 surgical video datasets: the Cholec80 dataset, the M2CAI dataset, and the Lobec100 dataset. The first two are benchmark datasets for surgical tool recognition in the literature, while the last one is a new dataset generated by ourselves.

\subsection{The Cholec80 Dataset}
\begin{figure}[H]
\centering
\setlength{\abovecaptionskip}{-1cm}
\includegraphics[width=\columnwidth]{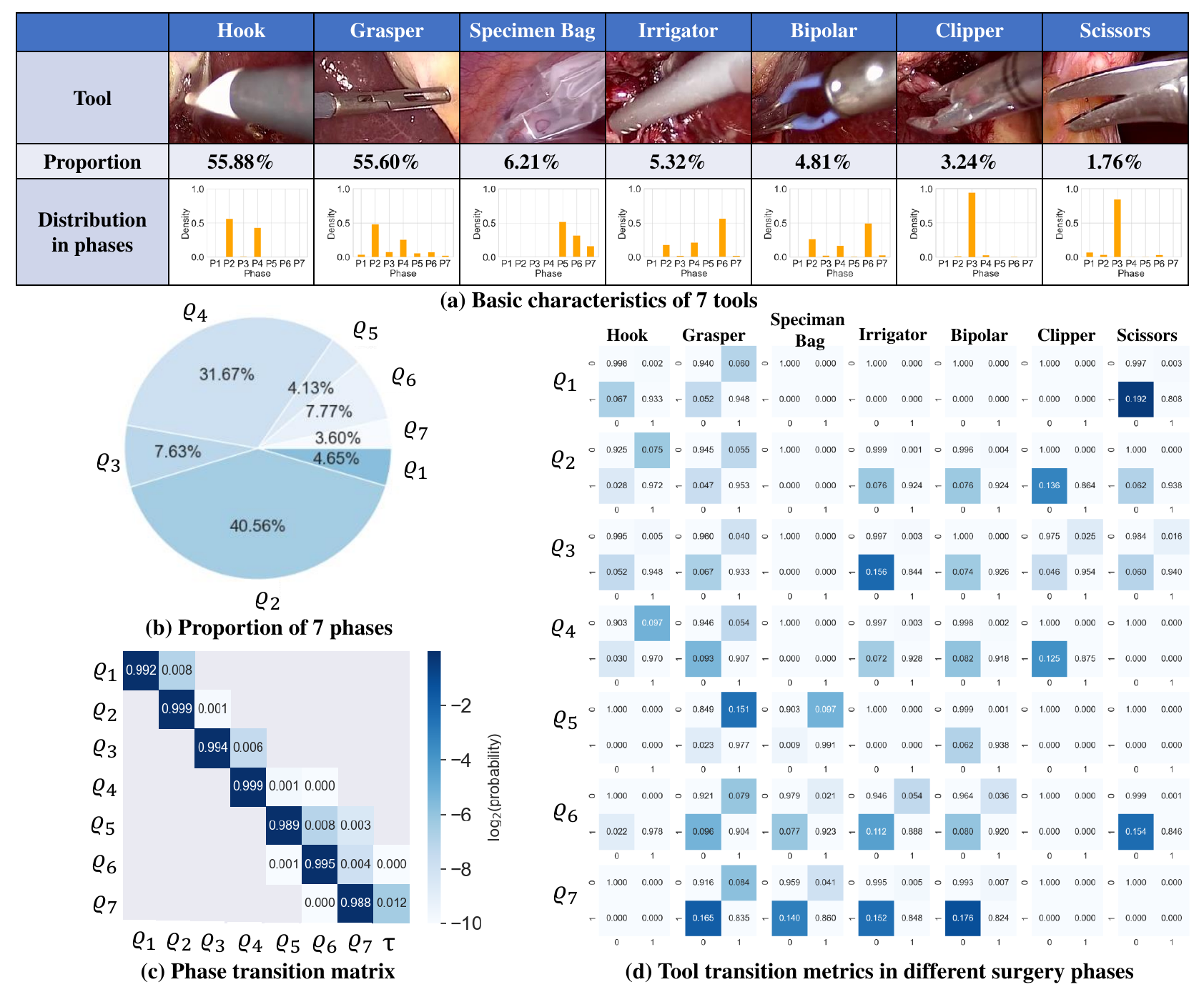}
\caption{Key features of the Cholec80 dataset. {\bf (a)} Basic summary on presence of the 7 major surgical tools in the Cholec80 dataset. 
{\bf (b)} Proportion of the 7 phases in cholecystectomy surgery. {\bf (c)} Phase transition matrix. {\bf (d)} Status transition matrices of 7 tools in 7 different surgical phases. (We only colored the anti-diagonal of the transition matrix.)
}
\label{fig:Cholec80}
\end{figure}

Containing videos of 80 cholecystectomy surgeries conducted in University Hospital of Strasbourg in 2016,
the Cholec80 dataset established by \cite{endonet} is a benchmark dataset for surgical tool recognition. 
The average duration of the 80 videos in this dataset is 38.4 minutes.
All videos were recorded at the resolution of $854 \times 480$ pixels with a shooting frequency of 25 fps, and were down-sampled into a lower frequency of 1 fps to generate the sequence of key frames for downstream analysis. 
For each key frame of this dataset, annotations for presence of 7 major surgical tools, including hook, grasper, specimen bag, irrigator, bipolar, clipper and scissors, and 7 primary surgical phases, including reparation ($\varrho_1$), Calot triangle dissection ($\varrho_2$), clipping and cutting ($\varrho_3$), gallbladder dissection ($\varrho_4$), gallbladder packaging ($\varrho_5$), cleaning and coagulation ($\varrho_6$) and gallbladder retraction ($\varrho_7$), were manually annotated by domain experts.
It took domain experts about 110 labor hours to manually generate these annotations.
Figure~\ref{fig:Cholec80} summarizes the key information about this dataset.


\subsection{The M2CAI Dataset}
Containing videos of 15 cholecystectomy surgeries with annotations on location of surgical tools, the M2CAI dataset established by \cite{frcnntool} is a benchmark dataset for surgical tool recognition and localization.
The original videos were recorded in University Hospital of Strasbourg in the same way as in the Cholec80 dataset, and were down-sampled to a sequence of key frames at the sampling frequency of 1 fps for downstream analysis as well.
Presence of the same set of 7 tools as in Cholec80 dataset were annotated for key frames of the 15 videos.
All key frames with just one tool, with a collection of additional key frames with more than one tool, were picked up for further annotation on the location of tools appearing in them, in term of rectangle bounding boxes.
About 90\% of key frames were dropped out in this procedure to reduce the annotation cost, with 10\% of key frames were annotated with bounding boxes of surgical tools, leading to an incomplete sequence of key frames from the original videos (with bounding boxes) to form the dataset.
Average length of the key frame sequences generated from the 15 surgical videos is 187.

\begin{figure}[H]
\centering
\setlength{\abovecaptionskip}{-1cm}
\includegraphics[width=\columnwidth]{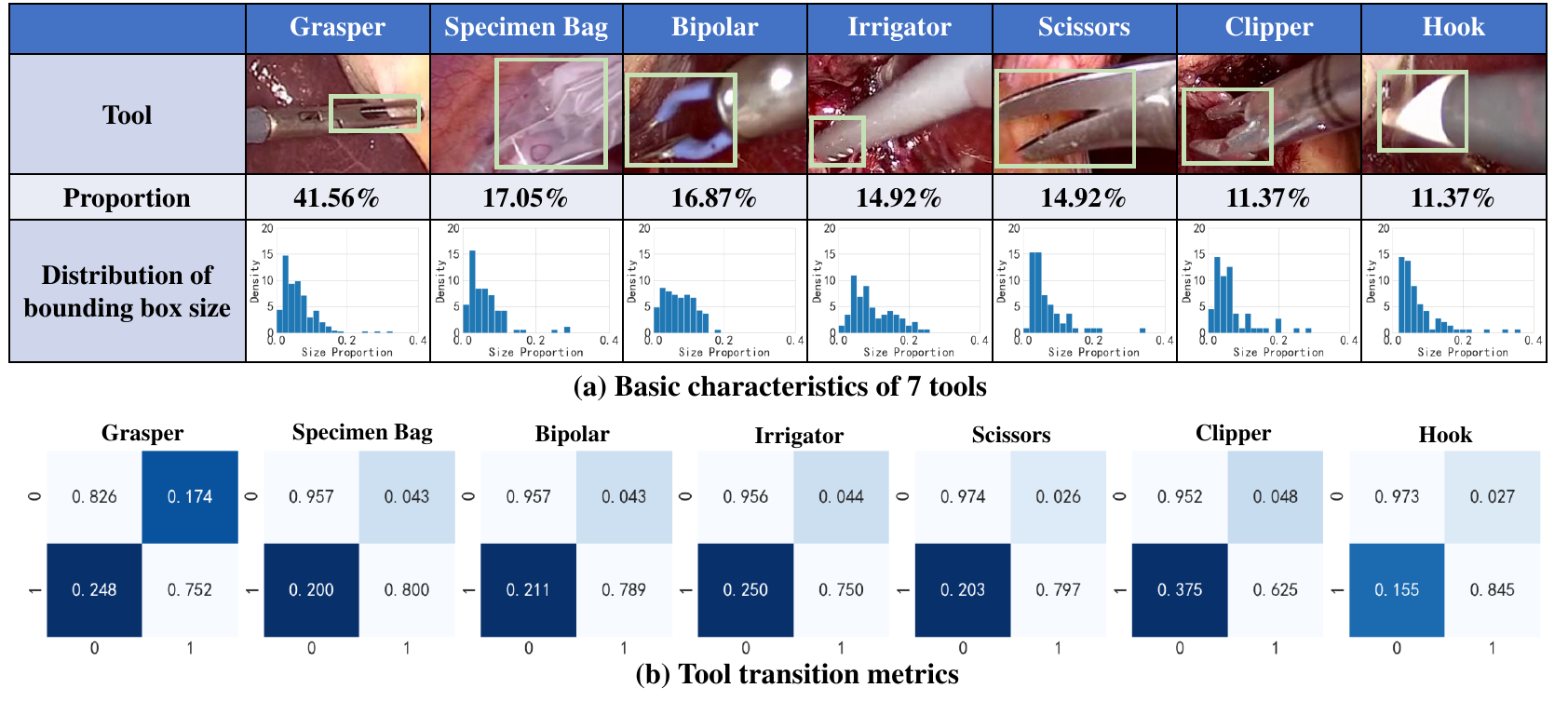}
\caption{Key features of the M2CAI dataset. {\bf (a)} Basic characteristics of the 7 tools of interest with bounding boxes highlighted. {\bf (b)} Tool transition matrices of 7 tools.}
\label{fig:M2CAI}
\end{figure}

As one of the earliest datasets with annotations on locations of surgical tools, the M2CAI dataset is a benchmark for the task of surgical tool recognition and localization. 
Please note that different from key frames in the Cholec80 dataset which are sampled with equal time gaps, key frames selected in the M2CAI dataset are not evenly distributed along the timeline.
Figure~\ref{fig:M2CAI} summarizes the key information about the M2CAI dataset.

\subsection{The Lobec100 Dataset}

The Cholec80 and M2CAI datasets are about cholecystectomy surgery. 
To expand the scope of surgeries, a new dataset called Lobec100 was introduced in this study. 
It contains videos of 100 lobectomy surgeries conducted in Peking University People’s Hospital in 2020 and 2021. 
The videos were recorded at two resolution levels ($1920 \times 1080$ or $720 \times 576$) and two shooting frequencies (125 fps or 25 fps), and were down-sampled into a lower frequency of 1 fps for downstream analysis. 
Lobectomy surgery is associated with longer duration, more types of surgical tools, and more surgical phases than cholecystectomy surgery.
The average duration of the Lobec100 videos is 108.7 minutes, which is 3 times longer than the average duration of the Cholec80 videos.
More than 20 different types of surgical tools appear in the Lobec100 videos, of which 10 are the major ones.

Because the Lobec100 videos are associated with longer duration and more surgical tools, it is more expensive to annotate them.
It took us more than 148 labor hours to fully annotate the presence labels of the 10 major tools in every key frame of 20 randomly selected videos. 
And, we would expect more than 1,000 labor hours for fully annotating all 100 videos in the Lobec100 dataset, which is apparently too expensive to implement in practice.
To obtain more effective training data at an affordable annotation cost, we trained a deep-learning-based recognizer for the 10 types of surgical tools with the 20 fully annotated videos only, and utilized the roughly trained recognizer as a filter to screen all key frames of the 80 unlabeled videos for potential candidates that contain the target tools.
Based on these selected candidates, we manually picked up for each tool 1,000 additional frames that contain the tool as additional positive samples for downstream analysis.
Finally, we came up with the current Lobec100 dataset composed of 3 types of videos: 20 fully labelled videos with all key frames manually labelled, 36 partially labelled videos with some key frames manually labelled, and 44 unlabeled videos whose key frames were not labelled at all.

\begin{figure}[H]
\centering
\setlength{\abovecaptionskip}{-1cm}
\includegraphics[width=\columnwidth]{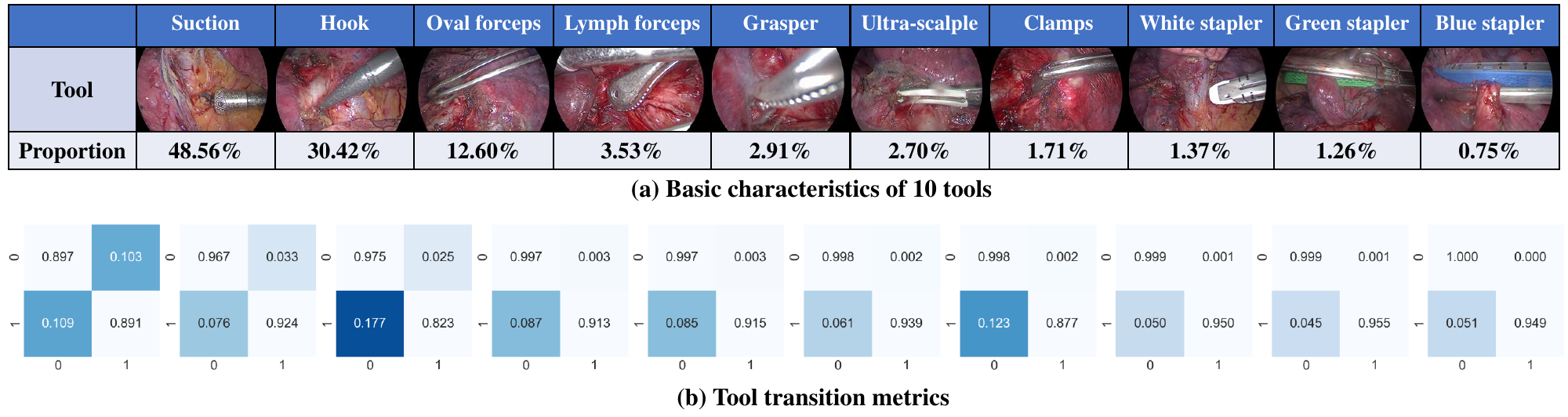}
\caption{Key features of the Lobec100 dataset. {\bf (a)} Basic characteristics of the 10 tools of interest. {\bf (b)} Tool transition matrices of 10 tools.}
\label{fig:Lobec}
\end{figure}

Compared to the traditional strategy to prepare annotations that annotates all key frames of the involved videos purely manually, as in Cholec80 and M2CAI, the algorithm-aided annotation strategy we used to establish the Lobec100 dataset enjoys lower annotation cost and more flexible annotation form, and better represents those less commonly used tools.
As we will show in detail in later sections, combined with the HMM-stabilized deep learning approaches proposed in this study, such an annotation strategy would lead to significant performance improvement with little additional annotation cost.
Appearance frequency and state transition probabilities of the 10 major tools in the 100 lobectomy surgeries were estimated based on the 15 fully annotated videos and illustrated in Figure~\ref{fig:Lobec}.

\subsection{Unique Characteristics of Surgical Videos}
Based on the above exploratory analyses, we found the following unique characteristics of the involved datasets. 
First, associated with a limited semantic space involving only a few surgical phases and a couple of surgical tools with no obvious long-distance correlation, the semantic structure of a surgical video is rather simple compared to general videos with an open set of objects and complex interactions.
Second, a surgery is usually conducted from one surgical phase (e.g., $\varrho_l$) to the next surgical phase (e.g., $\varrho_{l+1}$) in a nearly deterministic order with few exceptions.
Third, once a surgical tool appears or disappears in a frame, it tends to remain the same for a number of frames in the future, with transition between presence and absence a rare event in a surgical video.
Forth, different surgical phases are associated with different signature tools and the transition between presence and absence of a surgical tool follows heterogeneous statistical rules across different surgical phases.

These facts reveal the following insights immediately.
First, a low-order Markov chain would be sufficient to model the status transition patterns of surgical videos in most cases.
Second, it is necessary to build heterogeneous Markov chains for different surgical phases to reflect the phase-specific properties of a surgical video.
Third, dependence across different tools is secondary, when the status of surgical phase is given.
These insights suggest that a simple Markov model might be more appropriate for surgical tool recognition than complex deep leaning models.
To best of our knowledge, however, we did not find any attempts in the literature to implement and evaluate such an idea yet. 
In the later part of this paper, we will fill in the gap with explicit statistical models and concrete experimental evidences.

\section{HMM-Stabilized Deep Learning 
}\label{sec:HMM}
\subsection{The Statistical Model}\label{sec:HMM-model}
In a typical TPD problem, we focus on detecting whether an arbitrary image from a set of surgery videos contains specific types of surgical tools. 
Formally, suppose that a collection of $m$ surgical videos $\cV=\{\bV_i\}_{1\leq i\leq m}$ are under study, where the $i$-th video $\bV_i=(F_{i,t})_{1\leq t\leq n_i}$ are composed of $n_i$ key frames, and a set of $K$ surgical tools $\cT=\{\tau_k\}_{1\leq k\leq K}$ are of our interest.
Additionally, suppose that the entire surgical procedure is composed of multiple phases $\cP=\{\varrho_l\}_{1\leq l\leq L}$, where each phase $\varrho_l$ corresponds to a collection of specific actions and is often associated with some unique surgical tools.
Let $\bP_{i}=(P_{i,t})_{1\leq t\leq n_i}$ be the \emph{phase indicator vector} of video $V_i$, where $P_{i,t}=\varrho$ if key frame $F_{i,t}$ belongs to phase $\varrho$.
Let $\bT_{i,\tau}=(T_{i,t,\tau})_{1\leq t\leq n_i}$ be the \emph{tool presence vector} of tool $\tau$ in video $\bV_i$, where $T_{i,t,\tau}=1$ if tool $\tau$ presents in key frame $F_{i,t}$, and 0 otherwise.
Our primary goal in a TPD problem is to predict $\bT_i=\{\bT_{i,\tau}\}_{\tau\in\cT}$, sometimes $\bP_{i}$ as well, based on key frames in video $\bV_i$.


For a deep-learning-based TPD method $\cM$ that is ready to use in practice, let $\hat\bT_{i}=\{\hat\bT_{i,\tau}\}_{\tau\in\cT}$ and $\hat\bP_i$ be its predicted tool presence vectors and phase indicator vector for video $\bV_i$, where $\hat\bT_{i,\tau}=(\hat T_{i,t,\tau})_{1\leq t\leq n_i}$ and $\hat\bP_{i}=(\hat P_{i,t})_{1\leq t \leq n_i}$ with $\hat T_{i,t,\tau}$ and $\hat P_{i,t}$ being the element-wise predictors.
Apparently, the predicted values for key frame $F_{i,t}$, i.e.,  $\bY_{i,t}=(\hat P_{i,t},\{\hat T_{i,t,\tau}\}_{\tau\in\cT})$, is a noised version of the true status $\bX_{i,t}=(P_{i,t},\{T_{i,t,\tau}\}_{\tau\in\cT})$ of key frame $F_{i,t}$, because TPD method $\cM$ may make mistakes from time to time in prediction.
Define $\cI_i=(\bP_i,\bT_i;\hat\bP_i,\hat\bT_i)$ as the \emph{complete characteristic profile} of video $\bV_i$ under TPD method $\cM$, and denote $\cI=\{\cI_i\}_{1\leq i\leq m}$.
In reality, the dependence structure of elements in $\cI$ could be rather complicated, depending on both the intrinsic logic of the surgery process and the algorithmic structure of method $\cM$. 
Here, we chose to simplify the problem by assuming that $\cI$ follows 
the following joint likelihood:
\begin{align}
\label{eq:NHMM-1}
\bbP(\cI)&=\prod_{i=1}^m\bbP(\cI_i),\\
\label{eq:NHMM-2}
\bbP(\cI_i)&=\bbP\big(\bP_i,\bT_i;\hat\bP_i,\hat\bT_i\big)
= \bbP(\bP_i)\cdot\bbP(\bT_i\vert\bP_i)\cdot\bbP(\hat\bP_i\vert\bP_i)\cdot\bbP(\hat{\bT}_i\vert\bT_i);\\
\label{eq:NHMM-3}
\bbP(\bP_i)
&=\bbP(P_{i,1})\cdot\prod_{t=2}^{n_i}\bbP(P_{i,t}\vert P_{i,t-1})
=\textbf{Multinomial}(P_{i,1}\vert\balpha)\cdot\prod_{t=2}^{n_i} \bA(P_{i,t-1},P_{i,t}),\\
\label{eq:NHMM-4}
\bbP(\bT_i\vert\bP_i)&=\prod_{\tau\in\cT}\bbP(\bT_{i,\tau}\vert\bP_i)
=\prod_{\tau\in\cT}\left[\bbP(T_{i,1,\tau}\vert P_{i,1})\cdot\prod_{t=2}^{n_i}\bbP(T_{i,t,\tau}\vert T_{i,t-1,\tau};P_{i,t})\right]\\
&=\prod_{\tau\in\cT}\left[\textbf{Bernoulli}(T_{i,1,\tau}\vert\beta_{\tau,P_{i,1}})\cdot\prod_{t=2}^{n_i}\bA_{\tau,P_{i,t}}(T_{i,t-1,\tau},T_{i,t,\tau})\right];\nonumber\\
\label{eq:NHMM-5}
\bbP(\hat\bP_i\vert\bP_i)&=\prod_{t=1}^{n_i}\bbP(\hat P_{i,t}\vert P_{i,t})=\prod_{t=1}^{n_i} \bB(P_{i,t},\hat P_{i,t}),\\
\label{eq:NHMM-6}
\bbP(\hat{\bT}_i\vert\bT_i)&=\prod_{\tau\in\cT}\bbP(\hat\bT_{i,\tau}\vert\bT_{i,\tau})=\prod_{\tau\in\cT}\prod_{t=1}^{n_i}\bbP(\hat T_{i,t,\tau}\vert T_{i,t,\tau})=\prod_{\tau\in\cT}\prod_{t=1}^{n_i} \bB_{\tau}(T_{i,t,\tau},\hat T_{i,t,\tau});
\end{align}
where $\balpha=(\alpha_1,\cdots,\alpha_L)$ is the \emph{initial state distribution} of phase, $\bA=\{\bA(\varrho_i,\varrho_j)\}_{ \varrho_i,\varrho_j\in\cP}$ is the $L\times L$ \emph{state transition matrix} of the phase chains, $\bbeta=\{\beta_{\tau,\varrho}\}_{\tau\in\cT,\varrho\in\cP}$ is a $K\times L$ matrix with $\beta_{\tau,\varrho}$ being the \emph{initial presence probability} of the tool presence chain for tool $\tau$ in phase $\varrho$, $\bA_{\tau,\varrho}=\{\bA_{\tau,\varrho}(i,j)\}_{i,j \in \{0,1\}}$ is the $2\times 2$ \emph{state transition matrix} of the tool presence chains for tool $\tau$ in phase $\varrho$, $\bB=\{\bB(\varrho_i,\varrho_j)\}_{\varrho_i,\varrho_j\in\cP}$ is the $L\times L$ \emph{confusion matrix} of the phase chains, $\bB_\tau=\{\bB_\tau(i,j)\}_{i,j \in \{0,1\}}$ is the $2\times 2$ \emph{confusion matrix} of the tool presence chains for tool $\tau$.
Define $\cA=\{\bA_{\tau,\varrho}\}_{\tau\in\cT,\varrho\in\cP}$ and $\cB=\{\bB_\tau\}_{\tau\in\cT}$.
We have $\btheta=(\balpha,\bA;\bbeta,\cA;\bB,\cB)$ as the parameters of the model.
We refer to the above HMM-stabilized version of base method $\cM$ as $\cM_H$.
Compared to the frame-dependence method that models frame dependence via deep learning, i.e., $\cM_L$ and $\cM_A$, $\cM_H$ enjoys straightforward interpretation because its model parameters $\btheta$ encodes key medical insights about the surgery procedure and performance characteristics of method $\cM$.

\begin{figure}[!h]
\centering
\setlength{\abovecaptionskip}{-1cm}
\includegraphics[height=3.4in,width=\columnwidth]{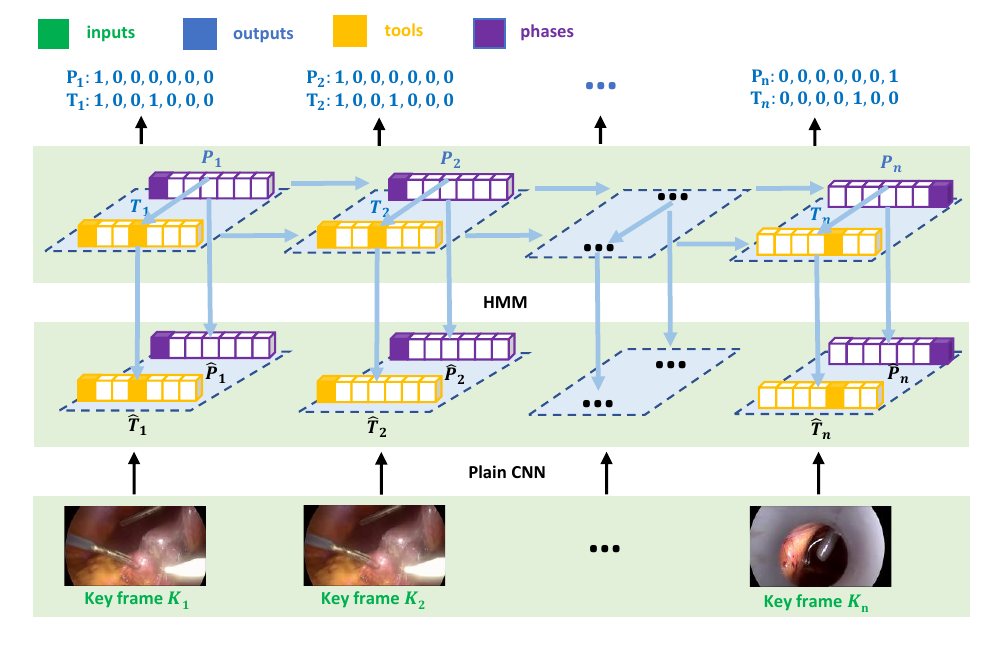}
\caption{A graphical illustration of the HMM-stabilized deep learning method for tool presence detection and phase recognition.}
\label{fig:NHMM-CNN}
\end{figure}

\subsection{Parameter Estimation via Semi-Supervised Learning}
If all elements of $\cI=\{\cI_i\}_{1\leq i\leq m}$ are completely observed, it would be straightforward to estimate $\btheta$. 
In practice, however, for a key frame $F_{i,t}$, although its predicted status $\bY_{i,t}=(\hat P_{i,t},\{\hat T_{i,t,k}\}_{1\leq k\leq K})$ is always observed as the output of TPD method $\cM$, its true status $\bX_{i,t}=(P_{i,t},\{T_{i,t,k}\}_{1\leq k\leq K})$ is typically unknown unless $F_{i,t}$ belongs to the training set of which every key frame has been manually labeled.
When some $\bX_{i,t}$'s are unobserved missing values, we would like to use the Expectation-Maximization (EM) algorithm \citep{EM} to estimate $\btheta$.

Denote $\bY_i=(\bY_{i,t})_{1\leq t\leq n_i}$ and $\bX_i^{obs}=\{\bX_{i,t}\}_{t\in\cL_i}$, where $\cL_i$ stands for the frames in video $\bV_i$ whose true status have been manually labelled.
Notation $\cI^{obs}_i=(\bX_i^{obs},\bY_i)$ highlights the observed elements of $\cI_i$.
Let $\cI^{obs}=\{\cI^{obs}_i\}_{1\leq i\leq m}$, and $\btheta^{(s)}$ be the estimation of $\btheta$ at the $s$-th step of the EM algorithm.
The E-step of the EM algorithm focuses on the calculation of the following $Q$-function:
\begin{eqnarray}
\label{NHMMQfunc}
Q(\btheta\vert\btheta^{(s)})&=&\bbE\left(\log\bbP(\cI)\vert \cI^{obs},\btheta^{(s)}\right)\nonumber\\
&=&\sum_{i=1}^m\bbE\left(\log\bbP(\cI_i)\vert \cI^{obs}_i,\btheta^{(s)}\right)
=\sum_{i=1}^m\sum_{\cI_i}\log\bbP(\cI_i)\bbP\left(\cI_i\vert \cI^{obs}_i,\btheta^{(s)}\right),
\end{eqnarray}
where
\begin{equation}
\bbP\left(\cI_i\vert \cI^{obs}_i,\btheta^{(s)}\right)=\frac{\bbP(\cI_i\vert\btheta^{(s)})\cdot \bbI(\cI_i\sim\cI_i^{obs})}{\sum_{\cI_i\sim\cI_i^{obs}}\bbP(\cI_i\vert\btheta^{(s)})},
\end{equation}
with notation $\cI_i\sim\cI_i^{obs}$ stands for the event that $\cI_i$ is consistent to $\cI_i^{obs}$, that all elements of $\cI_i^{obs}$ match  to their counterpart in $\cI_i$ exactly.

Maximizing $Q(\btheta\vert\btheta^{(s)})$ with respect to $\btheta$, we come up with the following updating function to iteratively improve the estiamtion of $\btheta$:
\begin{align}
\label{eq:Mstep1}
\alpha_\varrho^{(s+1)}&=\frac{\bbE\left[\bbN(P_{\cdot,1}=\varrho\vert \btheta^{(s)})\right]}{\sum_{\varrho'=1}^{L}\bbE\left[\bbN(P_{\cdot,1}=\varrho'\vert \btheta^{(s)})\right]},\\
\label{eq:Mstep2}
\beta_{\tau,\varrho}^{(s+1)}&=\frac{\bbE\left[\bbN(T_{\cdot,1,\tau}=1,P_{1}=\varrho\vert \btheta^{(s)})\right]}{\sum_{j'=0}^1\bbE\left[\bbN(T_{\cdot,1,\tau}=j',P_{\cdot,1}=\varrho\vert \btheta^{(s)})\right]},\\
\label{eq:Mstep3}
A^{(s+1)}(\varrho_i,\varrho_j)&=\frac{\bbE\left[\bbN(P_{\cdot,t-1}=\varrho_i,P_{\cdot,t}=\varrho_j\vert \btheta^{(s)})\right]}{\sum_{\varrho_{j'}=1}^{L}\bbE\left[\bbN(P_{\cdot,t-1}=\varrho_i,P_{\cdot,t}=\varrho_{j'}\vert \btheta^{(s)})\right]},\\
\label{eq:Mstep4}
B^{(s+1)}(\varrho_i,\varrho_j)&=\frac{\bbE\left[\bbN(P_{\cdot,t}=\varrho_i,\hat P_{\cdot,t}=\varrho_j\vert \btheta^{(s)})\right]}{\sum_{\varrho_{j'}=1}^{L}\bbE\left[\bbN(P_{\cdot,t}=\varrho_i,\hat P_{\cdot,t}=\varrho_{j'}\vert \btheta^{(s)})\right]},
\end{align}
\begin{align}
\label{eq:Mstep5}
A_{\tau,\varrho}^{(s+1)}(i,j)&=\frac{\bbE\left[\bbN(T_{\cdot,t-1,\tau}=i,T_{\cdot,t,\tau}=j,P_{\cdot,t}=\varrho\vert \btheta^{(s)})\right]}{\sum_{j'=0}^1\bbE\left[\bbN(T_{\cdot,t-1,\tau}=i,T_{\cdot,t,\tau}=j',P_{\cdot,t}=\varrho\vert \btheta^{(s)})\right]},\\
\label{eq:Mstep6}
B_\tau^{(s+1)}(i,j)&=\frac{\bbE\left[\bbN(T_{\cdot,t,\tau}=i,\hat T_{\cdot,t,\tau}=j\vert \btheta^{(s)})\right]}{\sum_{j'=0}^1\bbE\left[\bbN(T_{\cdot,t,\tau}=i,\hat T_{\cdot,t,\tau}=j'\vert \btheta^{(s)})\right]},
\end{align}
where
\begin{align}
\label{eq:Q-expecation}
\bbE\left[\bbN(P_{\cdot,1}=\varrho\vert \btheta^{(s)})\right] &= \sum_{i=1}^m\sum_{\cI_i}\bbI(P_{i,1}=\varrho)\bbP(\cI_i\vert\cI_i^{obs},\btheta^{(s)}),\\
\bbE\left[\bbN(T_{\cdot,1,\tau}=j,P_{\cdot,1}=\varrho\vert \btheta^{(s)})\right] &= \sum_{i=1}^m\sum_{\cI_i}\bbI(T_{i,1,\tau}=j,P_{i,1}=\varrho)\bbP(\cI_i\vert\cI_i^{obs},\btheta^{(s)}),\nonumber\\
\bbE\left[\bbN(P_{\cdot,t-1}=\varrho_i,P_{\cdot,t}=\varrho_j\vert \btheta^{(s)})\right] &= \sum_{i=1}^m\sum_{\cI_i}\bbI(P_{i,t-1}=\varrho_i,P_{i,t}=\varrho_j)\bbP(\cI_i\vert\cI_i^{obs},\btheta^{(s)}),\nonumber\\
\bbE\left[\bbN(P_{\cdot,t}=\varrho_i,\hat P_{\cdot,t}=\varrho_j\vert \btheta^{(s)})\right] &= \sum_{i=1}^m\sum_{\cI_i}\bbI(P_{i,t}=\varrho_i,\hat P_{i,t}=\varrho_j)\bbP(\cI_i\vert\cI_i^{obs},\btheta^{(s)}),\nonumber\\
\bbE\left[\bbN(T_{\cdot,t-1,\tau}=i,T_{\cdot,t,\tau}=j,P_{\cdot,t}=\varrho\vert \btheta^{(s)})\right] &= \sum_{i=1}^m\sum_{\cI_i}\bbI(T_{i,t-1,\tau}=i,T_{i,t,\tau}=j,P_{i,t}=\varrho)\bbP(\cI_i\vert\cI_i^{obs},\btheta^{(s)}),\nonumber\\
\bbE\left[\bbN(T_{\cdot,t,\tau}=i,\hat T_{\cdot,t,\tau}=j\vert \btheta^{(s)})\right] &= \sum_{i=1}^m\sum_{\cI_i}\bbI(T_{i,t,\tau}=i,\hat T_{i,t,\tau}=j)\bbP(\cI_i\vert\cI_i^{obs},\btheta^{(s)}).\nonumber
\end{align}

Direct calculation based on the above formula is clearly forbidden because it involves the enumeration of all possible values of $\cI_i$, whose complexity increases exponentially with the length of video $\bV_i$.
In practice, efficient computation with linear complexity can be achieved by following the standard Baum-Welch algorithm \citep{baumwelch}. 
We reserved these computational details to Appendix A.
Considering that the iterative estimation of the Baum-Welch algorithm is computationally expensive, we can also take a shortcut to estimate some of the parameters directly based on the training data only.
For example, the phase transition matrices $\bA$ and $\bB_{k,p}$'s can be conveniently estimated by the empirical transition matrices calculated from the observed phase labels in the training data.
The start probability $e_{k}$'s and $\bs$ can be estimated by the corresponding empirical frequencies.
Such a strategy is 
computationally convenient with little loss of estimation efficiency when videos with and without training labels have similar transition patterns.

\subsection{Inference of the Hidden States}
Given the estimated model parameters $\hat\btheta$, the unknown tool labels and phase labels $\bX_i=(\bP_i,\bT_i)$ in the testing video $i$ can be inferred from the predicted labels $\bY_i=(\hat\bP_i,\hat\bT_i)$ obtained from a TPD method $\cM$ by solving the following optimization problem:
\begin{equation}
\label{eq:HMM-stabilized methods prediction}
\tilde\bX_i=\arg\max_{\bX_i}\bbP\left(\bX_i\vert\bY_i;\hat\btheta\right),
\end{equation}
which can be resolved efficiently via a Viterbi-like algorithm \citep{viterbi}.
Alternatively, we also calculate
\begin{align}
    \dot \pi_{i,t,\tau}&=\sum_{\bP_i,\bT_i}\left[\bbP\left(\bP_i,\bT_i\vert\hat{\bP}_i,\hat{\bT}_i;\hat\btheta\right)\cdot \bbI(T_{i,t,\tau}=1)\right],\\
    \dot \lambda_{i,t,\varrho}&=\sum_{\bP_i,\bT_i}\left[\bbP\left(\bP_i,\bT_i\vert\hat{\bP}_i,\hat{\bT}_i;\hat\btheta\right)\cdot \bbI(P_{i,t}=\varrho)\right],
\end{align}
to summarize the overall probability to set $T_{i,t,\tau}=1$ and $P_{i,t}=\varrho$ after the HMM stabilization.
Meanwhile, we can also turn $\dot\pi_{i,t,\tau}$ into a binary indicator via thresholding and further obtain the precision-recall curve of HMM-stabilized methods for tool presence detection.
Efficient calculation of $\dot\pi_{i,t,\tau}$ and $\dot\lambda_{i,t,\varrho}$ can be achieved by dynamic programming.

\subsection{The Degenerated Cases}\label{subsec:DegeneratedHMM}
The general HMM-stabilized model defined in Section \ref{sec:HMM-model} involves a coupling of two HMMs, one for the phase track $(\bP_i,\hat\bP_i)$, and one for the tool track $(\bT_i,\hat\bT_i)$.
When only the phase track or the tool track is of interest, the general model would degenerate to a simplified model involving only one of the two original HMMs.

In case that phase recognition is not of our interest any more (as in ToolNet), where the phase indicator chain $\bP_i$ and the corresponding predictor $\hat\bP_i$ are not available for video $V_i$, the previous HMM degenerates to an ordinary HMM for the tool presence chain only with the following simplified likelihood:
\begin{align}
\label{eq:HMM-stabilized tools}
\bbP(\cI_i)&=\bbP\big(\bT_i,\hat\bT_i\big)=\bbP(\bT_i)\cdot\bbP(\hat{\bT}_i\vert\bT_i)=\prod_{\tau\in\cT}\Big[\bbP(\bT_{i,\tau})\cdot\bbP(\hat\bT_{i,\tau}\vert\bT_{i,\tau})\Big],\\
\bbP(\bT_{i,\tau})&=\bbP(T_{i,1,\tau})\cdot\prod_{t=2}^{n_i}\bbP(T_{i,t,\tau}\vert T_{i,t-1,\tau})\nonumber\\
&=\textbf{Bernoulli}(T_{1,\tau}\vert\beta_{\tau})\cdot\prod_{t=2}^{n_i}\bA_{\tau}(T_{i,t-1,\tau},T_{i,t,\tau}),\\
\bbP(\hat\bT_{i,\tau}\vert\bT_{i,\tau})&=\prod_{t=1}^{n_i}\bbP(\hat T_{i,t,\tau}\vert T_{i,t,\tau})=\prod_{t=1}^{n_i} \bB_{\tau}(T_{i,t,\tau},\hat T_{i,t,\tau}),
\end{align}
where $\bbeta=\{\beta_{\tau}\}_{\tau\in\cT}$ and $\bA_{\tau}=\{\bA_{\tau}(i,j)\}_{0\leq i,j\leq 1}$ are degenerated versions of $\bbeta$ and $\bA_{\tau,\varrho}$ in Section \ref{sec:HMM-model}, while $\bB_\tau$ keeps unchanged.
Parameter estimation and hidden state inference under the degenerated model can be achieved in the similar way with simplified formulation.
We reserve the detailed results to the Appendix.

On the other hand, in case that phase recognition, instead of tool recognition, is our primary interest, where tool presence indicators $\bT_i$ and the corresponding predictors $\hat\bT_i$ are not available, the general HMM would degenerate to another ordinary HMM for the phase indicator chain only with the following simplified likelihood:
{
\begin{align}
\label{eq:HMM-stabilized phases}
\bbP(\cI_i)&=\bbP\big(\bP_i,\hat\bP_i\big)=\bbP(\bP_i)\cdot\bbP(\hat{\bP}_i\vert\bP_i),\\
\bbP(\bP_i)
&=\bbP(P_{i,1})\cdot\prod_{t=2}^{n_i}\bbP(P_{i,t}\vert P_{i,t-1})\\
&=\textbf{Multinomial}(P_{i,1}\vert\balpha)\cdot\prod_{t=2}^{n_i} \bA(P_{i,t-1},P_{i,t}),\nonumber\\
\bbP(\hat\bP_i\vert\bP_i)&=\prod_{t=1}^{n_i}\bbP(\hat P_{i,t}\vert P_{i,t})=\prod_{t=1}^{n_i} \bB(P_{i,t},\hat P_{i,t}),
\end{align}
where $\balpha,\bA,\bB$ are the same as the parameters of HMM in Section \ref{sec:HMM-model}.
}


\subsection{Potential Extensions}
The proposed HMM-stabilized deep learning model can be further extended in a few directions.
First, considering that a TPD method $\cM$ based on deep learning typically makes probabilistic prediction on phase and tool presence for frame $F_{i,t}$, namely $\blambda_{i,t}=\{\lambda_{i,t,\varrho}\}_{\varrho\in\cP}$ and $\{\pi_{i,t,\tau}\}_{\tau\in\cT}$, behind the reported labels $\hat P_{i,t}$ and $\{\hat T_{i,t,\tau}\}_{\tau\in\cT}$, we can enhance the current discrete observations $\bY_{i,t}=(\hat P_{i,t},\{\hat T_{i,t,\tau}\}_{\tau\in\cT})$ for each frame by replacing $\bY_{i,t}$ with the underlying probabilities $\tilde\bY_{i,t}=(\blambda_{i,t},\{\pi_{i,t,\tau}\}_{\tau\in\cT})$.
As a multinomial distribution over the phase space $\cP$, $\blambda_{i,t}$ satisfies $0\leq \lambda_{i,t,\varrho}\leq 1$ and $\sum_{\varrho\in\cP}\lambda_{i,t,\varrho}=1$.
Encoding the predictive probability for event $T_{i,t,\tau}=1$, $\pi_{i,t,\tau}$ satisfies $0\leq \pi_{i,t,\tau}\leq 1$.
Typically, $\blambda_{i,t}$ and $\pi_{i,t,\tau}$ are output of the softmax activation function at the last layer of the corresponding deep modeling model.
By substituting $\bY_{i,t}=(\hat P_{i,t},\{\hat T_{i,t,\tau}\}_{\tau\in\cT})$ with $\tilde\bY_{i,t}=(\blambda_{i,t},\{\pi_{i,t,\tau}\}_{\tau\in\cT})$, we come up with an extended HMM with discrete latent space but numerical observations.
To work with the extended HMM, we need to update the simple emission models $\bbP(\hat P_{i,t}\vert P_{i,t})$ and $\bbP(\hat T_{i,t,\tau}\vert T_{i,t,\tau})$ in Eq.~\eqref{eq:NHMM-5} and \eqref{eq:NHMM-6} to more sophisticated ones, e.g.,
\begin{align}
\bbP(\blambda_{i,t}\vert P_{i,t}=\varrho)&=\textbf{Dirichlet}(\blambda_{i,t};\bphi_\varrho),\ \forall\ \varrho\in\cP;\nonumber\\
\bbP(\pi_{i,t,\tau}\vert T_{i,t,\tau}=\delta)&=\textbf{Beta}(\pi_{i,t,\tau};\bpsi_{\tau,\delta}),\ \forall\ \delta\in\{0,1\};\nonumber
\end{align}
where $\bphi_{\varrho}$ is the parameter vector of a $L$-dimensional Dirichlet distribution for modelling the output probability of $\blambda_{i,t}$ when $P_{i,t}=\varrho$, and $\bpsi_{\tau,\delta}$ is the parameter vector of a Beta distribution for modelling the output probability of $\pi_{i,t,\tau}$ when $T_{i,t,\tau}=\delta$.
Although this extended model involves more parameters and need more computation for parameter estimation and hidden state inference, there is no essential difficulty in implementing it.

Second, we can release the assumption of independent tool tracks in Eq.~\eqref{eq:NHMM-4} and \eqref{eq:NHMM-6} to allow interactions across different tool tracks.
For example, we can adopt the more general state transition model below
\begin{equation}\label{eq:NHMM-TranModel4Tools}
\bbP(\bT_{i}\vert \bP_i)=\prod_{\tau\in\cT}\left[\bbP(T_{i,1,\tau}\vert P_{i,1})\cdot\prod_{t=2}^{n_i}\bbP(T_{i,t,\tau}\vert\{T_{i,t-1,\tau}\}_{\tau\in\cT};P_{i,t})\right],
\end{equation}
where the status of $T_{i,t,\tau}$ depends on both $P_{i,t}$ and $\{T_{i,t-1,\tau}\}_{\tau\in\cT}$.
In practice, we can model $\bbP(T_{i,t,\tau}\vert\{T_{i,t-1,\tau}\}_{\tau\in\cT};P_{i,t}=\varrho)$ by a logistic regression with $\{T_{i,t-1,\tau}\}_{\tau\in\cT}$ as the predictors, $T_{i,t,\tau}$ as the response, and $\bm\eta_\varrho$ as the $\varrho$-specific regression coefficients.
On the other hand, we can also adopt the more general emission model below
\begin{equation}\label{eq:NHMM-EmissModel4Tools}
\bbP(\hat\bT_{i}\vert\bT_{i})=\prod_{\tau\in\cT}\left[\prod_{t=1}^{n_i}\bbP(\hat T_{i,t,\tau}\vert\{T_{i,t,\tau'}\}_{\tau'\in\cT})\right],
\end{equation}
where the status of $\hat T_{i,t,\tau}$ depends on the full status vector $\{T_{i,t,\tau'}\}_{\tau'\in\cT}$, instead of $T_{i,t,\tau}$ only.
Similarly, $\bbP(\hat T_{i,t,\tau}\vert\{T_{i,t,\tau'}\}_{\tau'\in\cT})$ can be modelled by a logistic regression with $\{T_{i,t,\tau'}\}_{\tau'\in\cT}$ as predictor and $\hat T_{i,t,\tau}$ as response as well.
The more general emission model in Eq.~\eqref{eq:NHMM-EmissModel4Tools} may be particularly useful when the target videos contain surgical tools of similar shapes that are difficult to distinguish by the backbone CNN.
In this case, the logistic regression model for $\bbP(\hat T_{i,t,\tau}\vert\{T_{i,t,\tau'}\}_{\tau'\in\cT})$ would give us quantitative insights about how appearance of one tool influences the recognition of another tool.

Third, in the current approach we fine tune the backbone CNN with the fully labelled videos only, and keep it fixed when stabilize its outputs with HMM.
In case that the HMM-stabilized deep learning approach can make reliable predictions about the unknown phase and tool labels for unlabeled or partially labelled videos, it may be a good idea to use these predicted labels as additional data to improve fine tuning of the backbone CNN, and feed the improved backbone CNN back to the system.
Such an iteration may help the HMM component and backbone CNN to enhance each other a little bit, and thus boost the performance of proposed HMM-stabilized deep learning further.

\section{Experimental Study}\label{sec:Experiments}
In this section, we evaluate the performance of the proposed HMM-stabilized deep learning methods with respect to existing methods via experimental study.
A wide range of experiments showed that the HMM-stabilized methods enjoy competitive performance, low computation cost, and flexible model training.
These results provide a strong empirical evidence to our insight that integrating deep learning and statistical learning wisely would lead to more powerful methods with good performance, transparent interpretation, and efficient utilization of domain knowledge.

\subsection{Study Design}\label{subsec:Study Design}
Here, we took 4 frame-independent methods, ToolNet, EndoNet, SwinNet, and FR-CNN, as the baseline methods for surgical tool recognition in different scenarios: ToolNet is for the scenario where only tool presence recognition is considered; EndoNet and SwinNet are for the scenario where tool presence recognition and phase recognition are considered jointly; and, FR-CNN is for the scenario where tool presence recognition is entangled with tool localization.
For each baseline method $\cM$, let $\cM_L$, $\cM_A$ and $\cM_H$ be its frame-dependent versions where the frame dependence is modeled by LSTM, attention mechanism, and HMM, respectively.
We would like to show by experiments that $\cM_H\approx\cM_A>\cM_L>\cM$ in terms of recognition accuracy for all 4 baseline methods.

To compare the performance of these competing methods systematically, we designed 3 groups of experiments.
The first group of experiments are under scenario $\bbS_C$ where continuous annotations for all key frames in a collection of surgical videos are available as training data.
The second group of experiments are under scenario $\bbS_D$ where $N_p$ positive frames and $N_n$ negative frames were selected for each tool to serve as training data for tool recognizers.
All selected positive frames contain the corresponding tool and are associated with annotations on tool presence, tool location and surgical phase, while the negative frames are obtained by randomly sampling frames in the surgical videos of interest and filtering out the positive ones.
Note that training frames generated in this way are discontinuous in sense that only some, not all, key frames are annotated for model training.
Compared to scenario $\bbS_C$, however, scenario $\bbS_D$ enjoys a much lower cost on preparing the data annotation.
The third group of experiments are under scenario $\bbS_{C+D}$, a combination of scenario $\bbS_{C}$ and scenario $\bbS_{D}$, where both continuous annotations for complete videos and discontinuous annotations for selective positive and negative frames are available for model training.
Table~\ref{tab:settings} summarizes the experimental settings for the 3 datasets Cholec80, M2CAI and Lobec100 under the 3 scenarios $\bbS_C$, $\bbS_D$ and $\bbS_{C+D}$.

\begin{table*}[h]
\renewcommand{\arraystretch}{0.7}
\renewcommand\cellgape{\Gape[3pt]}
\renewcommand\cellset{}
\setlength{\tabcolsep}{4pt}
\centering
\resizebox{\columnwidth}{!}{
\begin{tabular}{ccc c ccc c c}
\hline																									
	&		&		&&	\multicolumn{3}{c}{Training}							\\										
								\cline{5-7}																	
Dataset	&	Annotation	&	Methods supported	&&	$\bbS_{\cC}$	&	$\bbS_{\cD}$	&	$\bbS_{\cC+\cD}$	&&	Testing	\\										
\cline{1-3}								\cline{5-7}						\cline{9-9}											
Cholec80	&	\makecell{Tool presence \\ Surgical phase}	&	\makecell{ToolNet family \\ EndoNet family \\ SwinNet family}	&&	\makecell{86304 key frames \\ of the first 40\\ videos}	&	\makecell{1000 positive \& \\ negative samples\\ for each tool}	&	$-$	&&	\makecell{98194 key frames\\ of  the last 40\\ videos}	\\ \hline										
M2CAI		&	\makecell{Tool presence  \\ Tool location}	&	\makecell{ToolNet family \\ FR-CNN family \\ SwinNet family}	&&	\makecell{2248 selected\\ key frames}	&	\makecell{100 positive \& \\ negative samples\\ for each tool}	&	$-$	&&	\makecell{563 key frames\\ not selected in $\bbS_{\cC}$}	\\ \hline										
Lobec100	&	Tool presence	&	\makecell{ToolNet family \\ SwinNet family}	&&	\makecell{92071 key frames\\ of the first 15\\ videos}	&	\makecell{1000 positive \& \\ negative samples\\ for each tool}	&	\makecell{The union of\\ $\bbS_{\cC}$ and $\bbS_{\cD}$}	&&	\makecell{35696 key frames\\ of the other 5\\ videos}	\\ \hline															
\end{tabular}
}
\caption{Experimental settings for the 3 datasets Cholec80, M2CAI and Lobec100 under the 3 scenarios $\bbS_C$, $\bbS_D$ and $\bbS_{C+D}$.
}
\label{tab:settings}
\end{table*}

\subsection{Pre-training and training the involved deep learning models}
All methods considered in this study are composed of 3 key components: a pre-trained backbone network to extract image features from surgical video frames, an additional time series model to count for frame dependence (if involved), and a classification network (typically a fully connected layer) to classify frames based on the extracted features.
The FR-CNN family, however, contains an additional component for detecting the location of tools appearing in a frame. 
To implement any of these methods, we need to pick up an appropriate pre-trained backbone CNN, and conduct a model training procedure to fine-tune the initial parameters of the backbone CNN and train the model for frame dependence and the classification network with labelled surgical videos.

Table~\ref{tab:MethodSummary} summarizes the pre-trained backbone CNNs originally utilized by different approaches, including AlexNet \citep{alexnet}, VGGNet16 \citep{vgg}, ResNet18 and ResNet50 \citep{resnet}, and Swin-B \citep{swintransformer}.
All these backbone CNNs were pre-trained with the ImageNet dataset \citep{imagenet}, and can be downloaded from public platforms such as GitHub (\url{https://github.com/}).
In general, a pre-trained backbone CNN appeared later in literature enjoys a stronger ability on feature extraction from images at the price of heavier computation.
In this study, we implemented all methods of the ToolNet and EndoNet families with ResNet18 as the backbone, all methods of the FR-CNN family with ResNet50 as the backbone, and all methods of the SwinNet family with Swin-B as the backbone.

In the model training step, we compressed all key frames in the training dataset to the resolution level of $224\times 224$ to fit the input mode of the selected backbone CNNs. 
Following the widely adopted strategy for data augment \citep{vgg, resnet, swintransformer}, we randomly conducted a left-right symmetric transformation for each input video frame to ensure data diversity. 
All methods were trained by Adam optimizer \citep{adam} under the PyTorch framework.
Training of the SwinNet family was conducted on a Nvidia A100 GPU card of 80G graphical memory.
Training of all the other methods was conducted on a Nvidia RTX 3090 GPU card of 24G graphical memory. 
More details about the model training can be found from the code in Supplementary Materials.

\subsection{Measurements for Performance Evaluation}
Following \cite{endonet}, \cite{frcnntool} and \cite{mtrcnet},
we choose \emph{average precision} (AP) and \emph{mean average precision} (mAP) defined below as the primary measurements for performance evaluation.
Let $\cF$ be the collection of $m$ key frames in the testing videos, $T_{f,\tau}$ be the true presence label of tool $\tau$, $\pi_{f,\tau}$ be the predictive probability of tool $\tau$ to appear in a key frame $f\in\cF$ output by a surgical tool recognizer $\cM$.
For a given cutoff parameter $\lambda\in(0,1)$, the precision and recall of recognizer $\cM$ for recognizing tool $\tau$ under cutoff $\lambda$ are defined as:
\begin{align*}
    P_{\tau}(\lambda)&=\frac{\sum_{f\in\cF} \bbI(T_{f,\tau} = \bbI(\pi_{f,\tau} > \lambda) = 1)}{\sum_{f\in\cF} \bbI(\pi_{f,\tau} > \lambda)},\\
    R_{\tau}(\lambda)&=\frac{\sum_{f\in\cF} \bbI(T_{f,\tau} = \bbI(\pi_{f,\tau} > \lambda) = 1)}{\sum_{f\in\cF} \bbI(T_{f,\tau}=1)}.
\end{align*}
For any tool $\tau\in\cT$, the AP of recognizing $\tau$ by recognizer $\cM$, which is defined as the area under the corresponding precision-recall curve, can be calculated as follows:
\begin{equation}
   \text{AP}_\tau = \sum_{i=1}^m P_{\tau}(\lambda_{i-1,\tau})(R_{\tau}(\lambda_{i,\tau})-R_{\tau}(\lambda_{i-1,\tau})),
\end{equation}
where $\{\lambda_{i,\tau}\}_{1\leq i\leq m}$ are the ordered statistics of $\{\pi_{f,\tau}\}_{f\in\cF}$ with $\lambda_{0,\tau}=0$.
To evaluate the overall performance of recognizer $\cM$, we averaged the AP$_\tau$'s of $K$ tools to form mAP:
\begin{equation}
    \text{mAP} = \frac{1}{K}\sum_{\tau \in \cT}\text{AP}_\tau.
\end{equation}

Following \cite{mtrcnet}, we selected the \emph{F1-score} defined below as the primary metric for performance evaluation of phase recognition.
Let $\cF$ denote the set of $m$ key frames in the testing videos, $P_{f}$ and $\hat P_{f}$ represent the true label and predicted labels of phase about a key frame $f\in\cF$.
The precision and recall of a surgical phase classifier $\cM$ for classifying any phase $\varrho \in \cP$ are defined as follows:
\begin{align*}
    P_{\varrho}&=\frac{\sum_{f\in\cF}\bbI(P_{f}=\hat P_{f}=\varrho)}{\sum_{f\in\cF}\bbI(\hat P_{f}=\varrho)},\nonumber \\
    R_{\varrho}&=\frac{\sum_{f\in\cF}\bbI(P_{f}=\hat P_{f}=\varrho)}{\sum_{f\in\cF}\bbI(P_{f}=\varrho)}. \nonumber
\end{align*}
For any phase $\varrho\in\cP$, the F1-score of classifying $\varrho$ by recognizer $\cM$ is defined as:
\begin{equation}
   \text{F1}_{\varrho}=\frac{2\cdot P_{\varrho}\cdot R_{\varrho}}{P_{\varrho}+R_{\varrho}}.
\end{equation}
To calculate the overall performance of recognizer $\cM$ on all surgical phases, we averaged the F1-score$_\varrho$'s of $L$ phases as below:
\begin{equation}
    \text{mF1} = \frac{1}{L}\sum_{\tau \in \cT}\text{F1}_{\varrho}.
\end{equation}

\subsection{Performance on Surgical Tool and Phase Recognition}
Applying methods in ToolNet family, EndoNet family, FR-CNN family and SwinNet family
to the 3 datasets Cholec80, M2CAI and Lobec100 under the 3 scenarios $\bbS_C$, $\bbS_D$ and $\bbS_{C+D}$, and evaluating their performance on tool recognition (in terms of AP and mAP) and phase recognition (in terms of F1 and mF1), we come up with the summary table as showed in Table~\ref{tab:MainResults}.
From the table, we can see immediately that $\cM_H>\cM_L>\cM$ for any baseline method $\cM$ (i.e., ToolNet, EndoNet, FR-CNN or SwinNet) in terms of recognition accuracy under all settings.
These results suggest that the proposed HMM-stabilized deep learning is an effective strategy to enhance a baseline method for surgical tool recognition which outperforms the more sophisticated LSTM-enhanced version uniformly.

Because source codes of existing attention-based methods (i.e., ToolNet$_A$, EndoNet$_A$ and SwinNet$_A$) are not publicly available yet, we cannot directly apply these methods in the current experiment study.
Moreover, because the implementation of these methods involves many critical details that were not fully disclosed in the published papers, it is non-trivial to implement them by ourselves.
These facts make it infeasible to report detailed results of these attention-based methods in Table~\ref{tab:MainResults}.
However, according to the summarized results reported in the original papers, the tool recognition performance of ToolNet$_A$ and SwinNet$_A$ in Cholec80 under $\cS_C$ are mF1=71.5 and mAP=95.2, 
and the phase recognition performance of SwinNet$_A$ in Cholec80 under $\cS_C$ is accuracy=93.1 (results of EndoNet$_A$ are not formally reported in \cite{LAST}, and thus are not available yet).
Comparing these numbers to their counterpart obtained from ToolNet$_H$ and SwinNet$_H$, we are comfortable to claim that the proposed HMM-stabilized deep learning is still competitive even compared to most advanced deep learning architecture with attention mechanism.

\begin{figure}[t]
\centering
\setlength{\abovecaptionskip}{-1cm}
\includegraphics[width=\columnwidth]{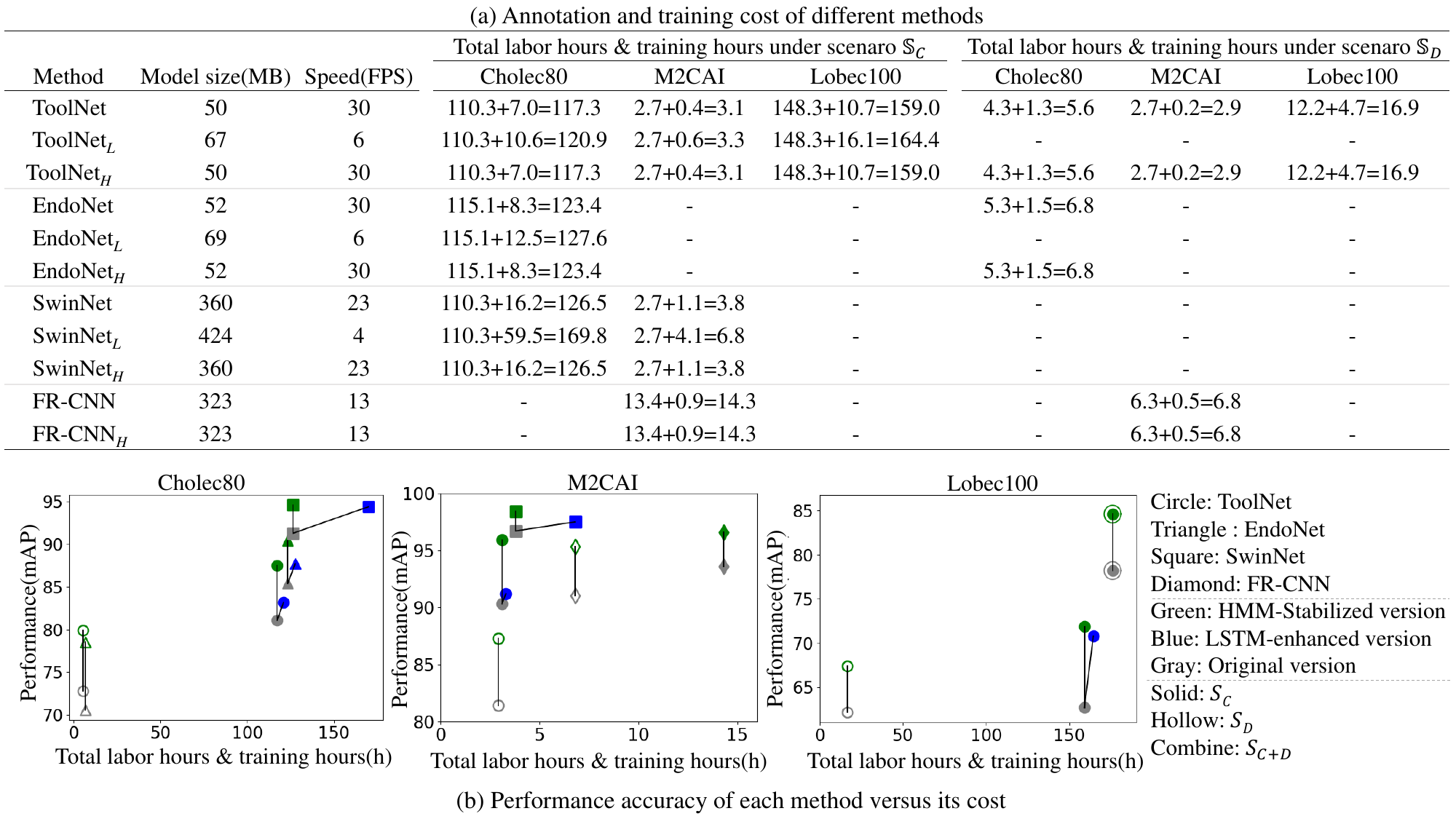}
\caption{Cost-Effectiveness analysis of different methods.}
\label{fig:cost}
\end{figure}

As showed in Table~\ref{tab:MainResults}, all existing frame-dependent methods that model frame dependence with LSTM or attention mechanism only work for the scenario with expensive continuous frame-level annotations, i.e., $\cS_C$.
The proposed HMM-stabilized deep learning, however, supports more flexible scenarios with cheap non-continuous annotations, e.g., $\bbS_D$ and $\bbS_{C+D}$.
Figure~\ref{fig:cost} (a) summarizes the data annotation and model training costs of various methods under $\cS_C$ and $\cS_D$, highlighting that the annotation cost in $\cS_D$ is indeed much cheaper than in $\cS_C$.
We note that although $\cM$ and $\cM_H$ encountered performance degeneration in $\cS_D$ with respect to $\cS_C$, their performance of in $\cS_{C+D}$ boosts significantly as showed in Table~\ref{tab:MainResults} (d).
This fact suggests that the flexible annotation in scenario $\cS_{C+D}$ has great potential to boost a tool recognition method with little annotation cost.

Figure~\ref{fig:cost} (b) visualizes the performance of each method on surgical tool recognition in term of mAP versus its cost in terms of labor hours for annotation and computation hours for model training in different dataset and experimental scenarios.
In these figures, each baseline method $\cM$ (gray) are connected to its LSTM-enhanced version $\cM_L$ (blue) and HMM-stabilized version $\cM_H$ (green) by two solid lines to highlight their ties.
From these figures we can see intuitively how different factors, including dataset, experimental scenario, baseline method and their extension, influence the performance, cost, and cost-effectiveness of a surgical tool recognition method.
All these results confirm explicitly that the proposed HMM-stabilized deep learning enjoys the best performance and the highest cost-effectiveness among all competing methods.
Particularly, the combination of the HMM-stabilized deep learning and the flexible annotation strategy $\cS_{C+D}$ may lead to a powerful tool for surgical tool recognition with low operating cost.

\begin{landscape}
\begin{table}[htbp]
\centering
\includegraphics[height=0.55\columnwidth]{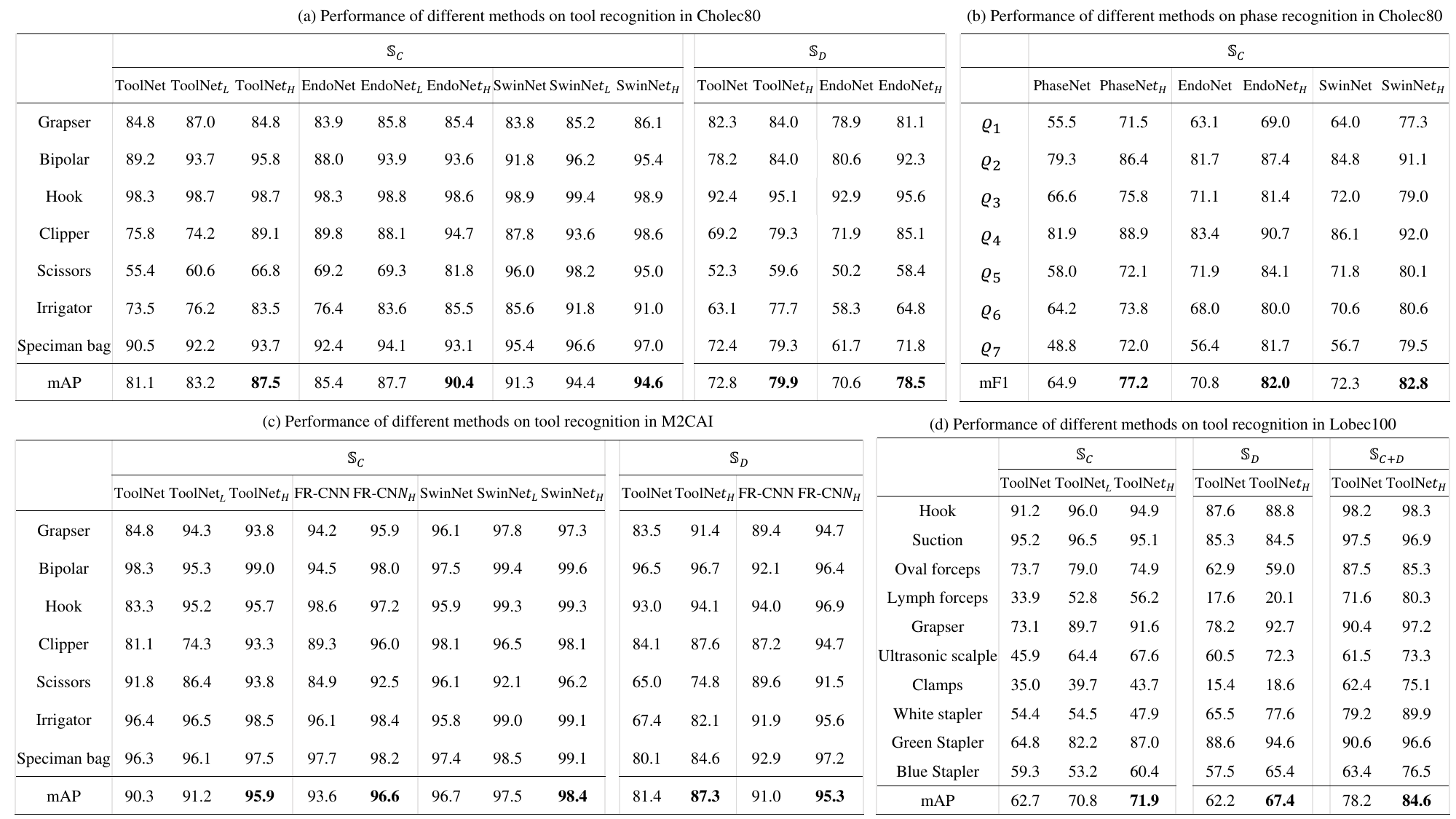}
\caption{Performance of different methods in tool presence detection and phase recognition.}
\label{tab:MainResults}
\end{table}
\end{landscape}

\section{Conclusions and Discussion}\label{sec:conclusion}
Modeling frame dependence of surgical videos by a classic HMM instead of more sophisticated deep learning architectures, such as LSTM or attention mechanism, we came up with a convenient framework to integrate statistical learning and deep learning for more effective recognition of surgical tools from surgical videos.
Compared to existing methods in literature purely based on deep learning, the proposed HMM-stabilized deep learning enjoys competitive performance, transparent interpretation, and high cost-effectiveness.
The success of HMM-stabilized deep learning in surgical videos analysis provides us a clear insight that over-flexible large deep learning models without a clear focus on the unique features of the problem of interest may lead to inefficient utilization of data and sub-optimal results.
Integrating statistical models and deep learning architectures wisely, however, often gives us a better solution to analysis of complex unstructured data.

The proposed approach can be further improved by replacing the discrete predicted labels for tool and phase presence with more accurate prediction probabilities, and/or introducing more complicated dependence structure in HMM to facilitate communication between various tool tracks.
We can also pursuit further improvement by feeding the HMM-stabilized predictions back into the original deep learning method as the extra pseudo training data.
Moreover, the algorithm-aided annotation strategy used to prepare the Lobec100 dataset has potential to serve as a general approach to reduce the annotation cost for preparing training data in this field.

\bibliographystyle{aaai}
\bibliography{myreference}

\newpage


\begin{appendices}
\appendix   

\setcounter{page}{1} 
\setcounter{table}{0}   
\setcounter{figure}{0}
\setcounter{section}{0}
\setcounter{equation}{0}
\renewcommand{\thetable}{S\arabic{table}}
\renewcommand{\thefigure}{S\arabic{figure}}
\renewcommand{\thesection}{S\arabic{section}}
\renewcommand{\theequation}{S\arabic{equation}}





\section{Deep Learning Models for Video Processing}
\subsection{Convolutional Neural Networks for Image Classification}
\begin{figure}[H]
\centering
\setlength{\abovecaptionskip}{-1cm}
\includegraphics[width=\columnwidth]{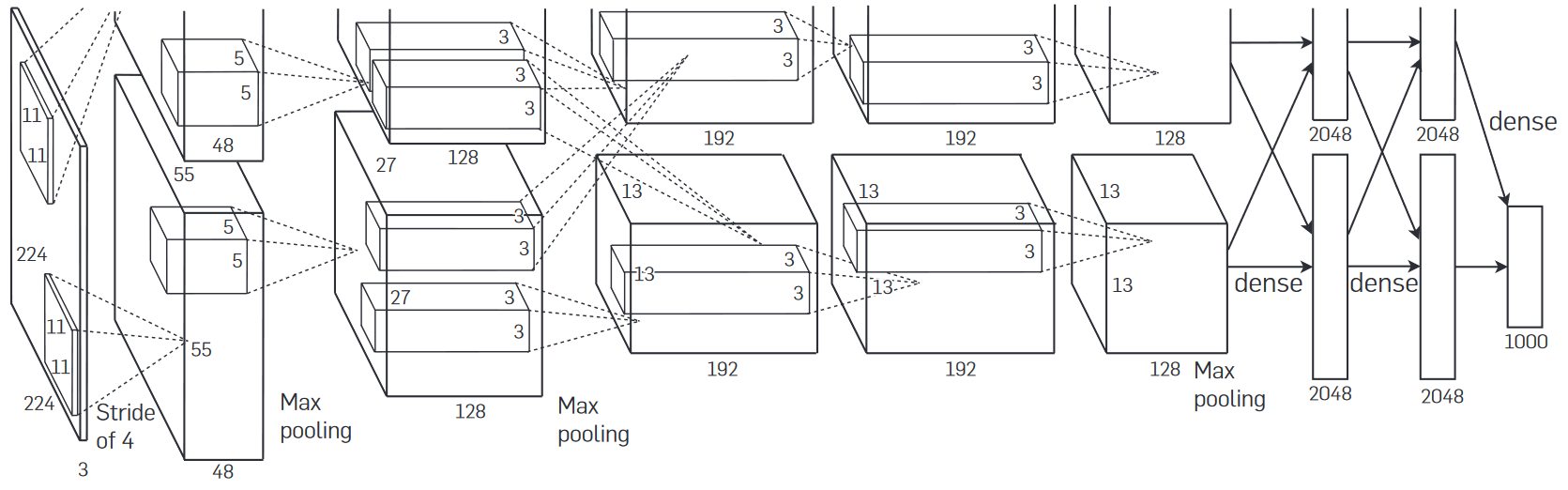}
\caption{The architecture of AlexNet.}
\label{fig:AlexNet}
\end{figure}

Proposed by \cite{firstcnn}, \emph{Convolutional Neural Network} (CNN) is a typical neural network for processing images. 
In 2012 Microsoft Image Classification Competition, a special variant of CNN named AlexNet \citepsm{alexnet} won the championship of the competition and exceeded the classification accuracy of traditional machine learning methods by 15\%, making people realize the great potential of CNN models in image processing. 
Inspired by the great success of AlexNet, more and more CNNs with 
various architectures were proposed in recent year, including VGG \citepsm{vgg}, GoogleNet \citepsm{googlenet}, ResNet \citepsm{resnet}, Vision-Transformer \citepsm{ViT}, Swin-Transformer \citepsm{swintransformer}. 
Figure~\ref{fig:AlexNet} displays the architecture of AlexNet \citepsm{alexnet}.

The good performance of CNNs mainly come from the following two points. 
First, a CNN utilizes the complex structure of neural networks and solves the problem of insufficient fitting in traditional machine learning models by providing a large number of parameters in image classification.
Secondly, compared to traditional fully connected neural networks, the CNN model uses exquisite convolutional kernel design to effectively extract local features of the image, preventing overfitting caused by too many parameters.

\subsection{Region-Based Convolutional Neural Networks for Object Detection}
Object detection refers to the image processing problem of highlighting locations of interest objects in images by horizontal rectangles exactly including the target objects. 
Each highlighted rectangle is usually called a \emph{Region of Interest} (RoI).
Traditionally, researchers often selected candidate regions and extracted the feature of candidate regions through artificially designed rules, and then used classic machine learning methods to find RoIs \citepsm{VJ,SIFT,HOG}.
These traditional methods have two fundamental limitations. 
First, the proposed candidate regions of each image are numerous and highly overlapped causing huge computational challenges for the RoI discovery. 
Second, the hand-crafted features are usually not flexible enough to represent the main feature of images and correctly choose the RoIs.

\begin{figure}[H]
\centering
\setlength{\abovecaptionskip}{-1cm}
\includegraphics[width=\columnwidth]{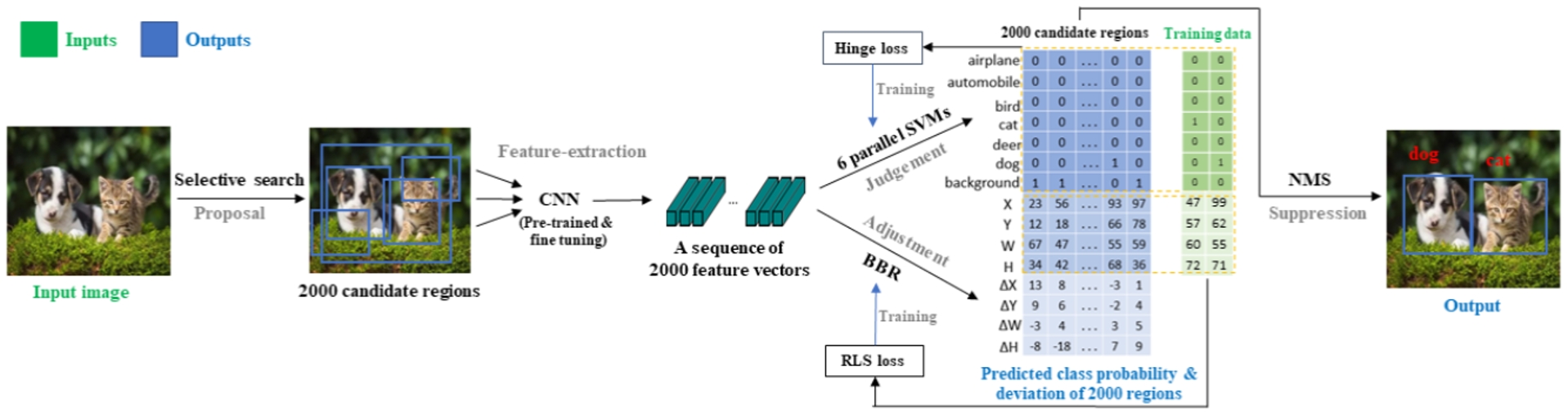}
\caption{The architecture of R-CNN.}
\label{fig:RCNN}
\end{figure}

Proposed by \citesm{rcnn}, \emph{Region-based Convolutional Neural Network} (R-CNN) is an important extension of CNN designed for object detection problem.
R-CNN incorporates deep learning into object detection problem and overcomes the two basic limitations of traditional methods by automatically training the selection function of candidate regions and the corresponding region features via CNN.
It consists of five main stages: a proposal stage that generates candidate regions for additional investigation; a feature-extraction stage that extracts important features for each proposed candidate region; a judgement stage that determines whether a candidate region contains an object of interest based on its feature vector; an adjustment stage that enhances the location and size of a candidate region; and a suppression stage that eliminates highly overlapping regions. 
Figure~\ref{fig:RCNN} shows the architecture of R-CNN \citepsm{ocr}.
Later on, many extensions of RNN have been proposed, including \emph{Faster R-CNN} \citepsm{fasterrcnn}, \emph{YOLO} \citepsm{yolo}, \emph{RetinaNet} \citepsm{retinanet}, \emph{EfficientDet} \citepsm{efficientdet}, \emph{DiffusionDet} \citepsm{diffusiondet}.

\subsection{RNN, LSTM and attention mechanism for processing sequence data}
\emph{Recurrent Neural Networks} (RNNs) refer to artificial neural networks characterized by a consistent architecture and layer flow. 
The Ising model, proposed by \citesm{Ising1} and \citesm{Ising2}, was the first RNN architecture without parameter learning capability.  
Hopfield networks are a kind of RNN that can be used to address optimization problems; they were first put out by \citesm{RNN1}.
\citesm{RNN2} introduced the back-propagation via time algorithm  for RNN training.
Since then, RNNs have been widely used for solving problems involving time series or sequence data, e.g., handwriting recognition, speech recognition, machine translation, robot control and so on, and gradually became the default choice for sequence data analysis in machine learning.

Proposed by \citesm{lstm}, \emph{Long Short-Term Memory} (LSTM)
is a variant of traditional RNN that can effectively capture the semantic association between long sequences and alleviate the phenomenon of gradient disappearance or explosion, with gate structures, including forget gate, input gate, cell state, and output gate. 
LSTM can better handle task involving time series or sequence data, because it solves the long-term dependency problem of RNN, and alleviates the ``gradient disappearance" problem caused by back propagation in training RNN.

\begin{figure}[H]
\centering
\setlength{\abovecaptionskip}{-1cm}
\includegraphics[height=2.8in,width=\columnwidth]{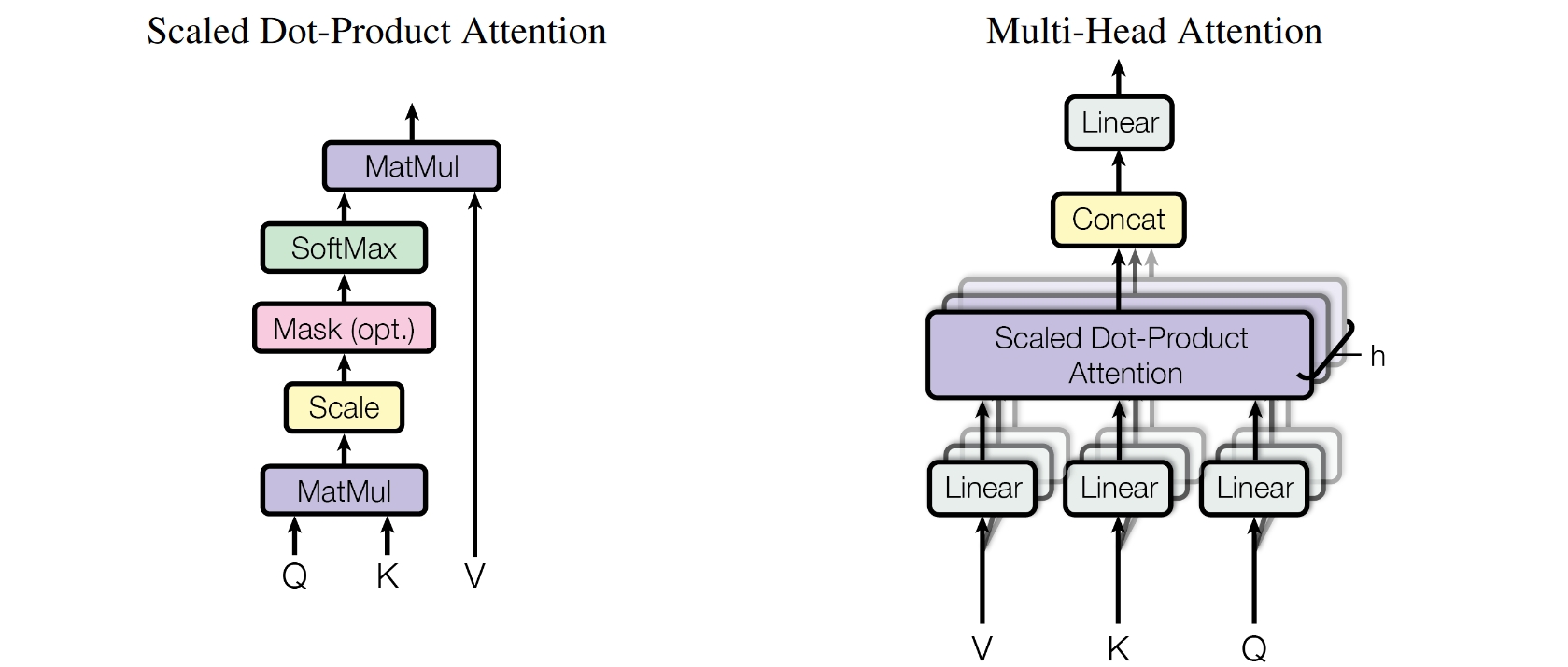}
\caption{The structure of attention mechanism.}
\label{fig:attention}
\end{figure}

Although LSTM is an effective tool to process sequence data, it is inefficient to model long-distance temporal correlation in a data sequence.
The \emph{attention mechanism}, however, is more powerful to handle long-distance temporal correlation by identifying the most relevant parts of the input sequences via a 2-dimensional attention map \citesm{attention}.
Figure~\ref{fig:attention} demonstrates the structure of attention mechanism.
One of the most popular applications of attention mechanism is in the transformer model, which has been widely used in natural language processing, including Bert \citepsm{bert}, GPT \citepsm{gpt}, and image processing, including Vision-Transformer \citepsm{ViT}, Swin-Transformer \citepsm{swintransformer}.

\subsection{Existing methods for surgical video analysis}
\begin{figure}[H]
\centering
\setlength{\abovecaptionskip}{-1cm}
\includegraphics[height=6in,width=\columnwidth]{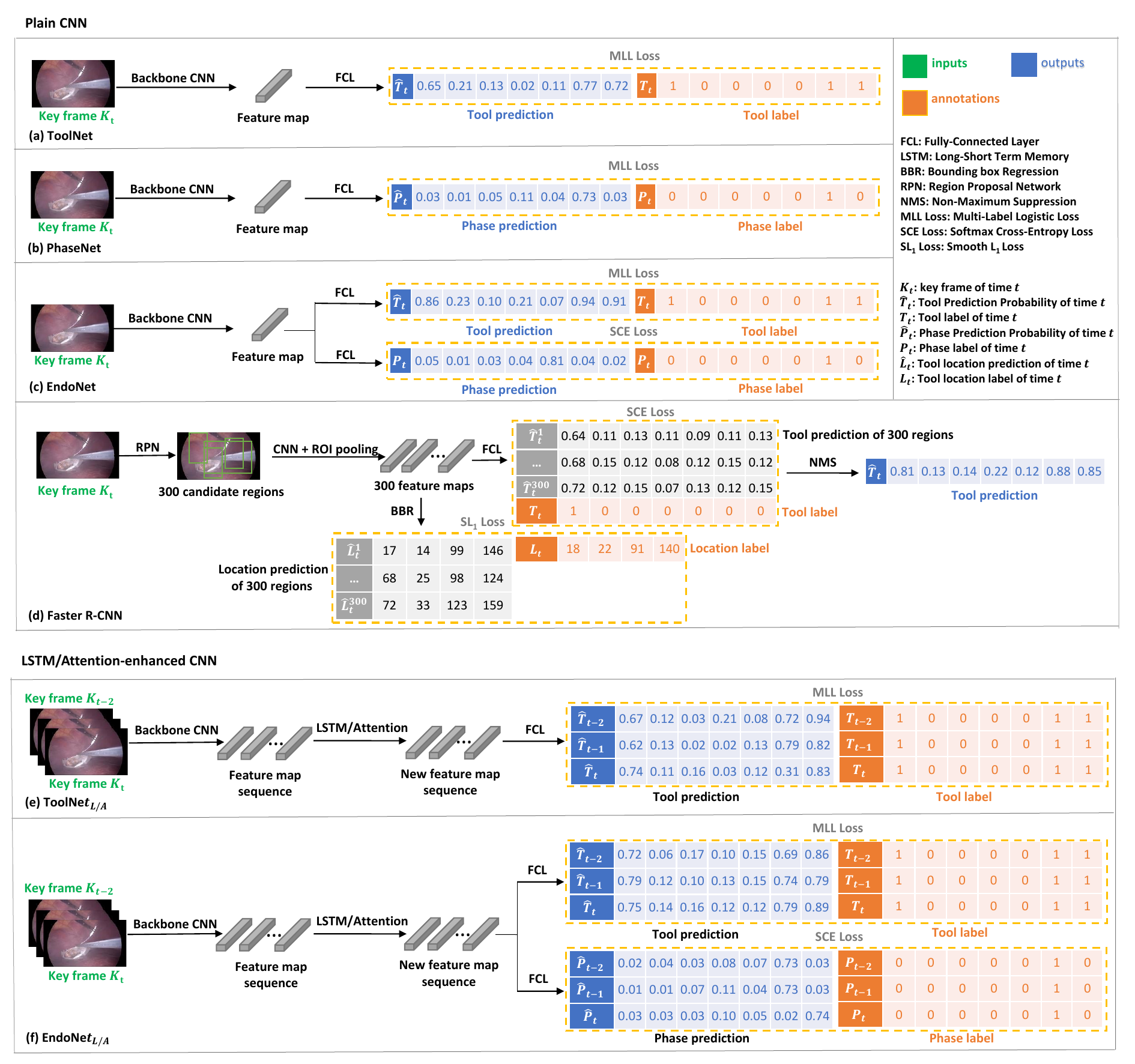}
\caption{Graphical illustration of deep-learning methods for surgical tool recognition.}
\label{fig:ExistingMethods}
\end{figure}

Figure~\ref{fig:ExistingMethods} shows the architecture of ToolNet, PhaseNet, EndoNet, SwinNet and some of their extensions equipped with LSTM and attention mechanism.

\section{Detailed Calculations for Inferring HMM-stabilized Deep Learning}
\subsection{Parameter Estimation via the EM Algorithm} \label{sec:HMM-EM}
In principle, the parameters of HMM-stabilized deep learning model in Section \ref{sec:HMM-model} can be estimated via the maximum likelihood principle, i.e., maximizing the likelihood defined in Eq.~\eqref{eq:NHMM-2}, which is a function of both true and predicted tool labels with respect to $\btheta$.
Because, true phase and tool labels are observed for training data only, we typically rely on the expectation-maximization algorithm \citepsm{EM} to do the optimization, treating the unobserved tool labels as missing data. 
The Q-function of E-step defined in Eq.~\eqref{NHMMQfunc} is shown as follows:
\begin{align}
Q(\btheta\vert\btheta^{(s)})=\sum_{i=1}^m\sum_{\cI_i}\log\bbP(\cI_i)\bbP\left(\cI_i\vert \cI^{obs}_i,\btheta^{(s)}\right),
\end{align}
where $\btheta^{(s)}$ is the estimation of model parameter $\btheta$ in the $s$-th iteration of the EM algorithm.

After substituting Eq.~\eqref{eq:NHMM-2} through \eqref{eq:NHMM-6} into the Q-function, we obtain
\begin{align}
\label{eqsm:NHMMDecomp}
Q(\btheta\vert\btheta^{(s)})=&\sum_{i=1}^m\sum_{\cI_i}\log\left(\textbf{Multinomial}(P_{i,1}\vert\balpha)\cdot\prod_{t=2}^{n_i} \bA(P_{i,t-1},P_{i,t})\right)\bbP\left(\cI_i\vert \cI^{obs}_i,\btheta^{(s)}\right) \nonumber\\
&+\sum_{i=1}^m\sum_{\cI_i}\log\left(\prod_{\tau\in\cT}\textbf{Bernoulli}(T_{i,1,\tau}\vert\beta_{\tau,P_{i,1}})\cdot\prod_{t=2}^{n_i}\bA_{\tau,P_{i,t}}(T_{i,t-1,\tau},T_{i,t,\tau})\right)\bbP\left(\cI_i\vert \cI^{obs}_i,\btheta^{(s)}\right) \nonumber \\
&+\sum_{i=1}^m\sum_{\cI_i}\log\left(\prod_{t=1}^{n_i} \bB(P_{i,t},\hat P_{i,t})\right)\bbP\left(\cI_i\vert \cI^{obs}_i,\btheta^{(s)}\right) \nonumber \\
&+\sum_{i=1}^m\sum_{\cI_i}\log\left(\prod_{\tau\in\cT}\prod_{t=1}^{n_i} \bB_{\tau}(T_{i,t,\tau},\hat T_{i,t,\tau})\right)\bbP\left(\cI_i\vert \cI^{obs}_i,\btheta^{(s)}\right).
\end{align}
By further grouping similar terms in Eq.~\eqref{eqsm:NHMMDecomp} based on the parameters $\btheta=(\balpha,\bA;\bbeta,\cA;\bB,\cB)$, we have
\begin{align}
\label{NHMMQfuncDecomp}
Q(\btheta\vert\btheta^{(s)})=&\sum_{\varrho\in\cP}\log\alpha_\varrho\bbE\left[\bbN(P_{\cdot,1}=\varrho\vert \btheta^{(s)})\right]\nonumber\\
&+\sum_{\varrho_i\in\cP}\sum_{\varrho_j\in\cP}\log \bA(\varrho_i,\varrho_j)\bbE\left[\bbN(P_{\cdot,t-1}=\varrho_i,P_{\cdot,t}=\varrho_j\vert \btheta^{(s)})\right]\nonumber \\
&+\sum_{\tau\in\cT}\sum_{l=0}^1((1-l)\log(1-\beta_{\tau,\varrho})+l\log(\beta_{\tau,\varrho}))\bbE\left[\bbN(T_{\cdot,1,\tau}=l,P_{\cdot,1}=\varrho\vert \btheta^{(s)})\right]\nonumber\\
&+\sum_{\tau\in\cT}\sum_{\varrho\in\cP}\sum_{m=0}^{1}\sum_{n=0}^{1} \log \bA_{\tau,\varrho}(m,n)\bbE\left[\bbN(T_{\cdot,t-1,\tau}=m,T_{\cdot,t,\tau}=n,P_{\cdot,t}=\varrho\vert \btheta^{(s)})\right]\} \nonumber \\
&+\sum_{\varrho_i\in\cP}\sum_{\varrho_j\in\cP}\log \bB(\varrho_i,\varrho_j)\bbE\left[\bbN(P_{\cdot,t}=\varrho_i,\hat P_{\cdot,t}=\varrho_j\vert \btheta^{(s)})\right] \nonumber \\
&+\sum_{\tau\in\cT}\sum_{m=0}^1\sum_{n=0}^1\log \bB_\tau(m,n)\bbE\left[\bbN(T_{\cdot,t,\tau}=m,\hat T_{\cdot,t,\tau}=n\vert \btheta^{(s)})\right],
\end{align}
where 
\begin{align}
\bbE\left[\bbN(P_{\cdot,1}=\varrho\vert \btheta^{(s)})\right] &= \sum_{i=1}^m\sum_{\cI_i}\bbI(P_{i,1}=\varrho)\bbP(\cI_i\vert\cI_i^{obs},\btheta^{(s)}),\nonumber\\
\bbE\left[\bbN(T_{\cdot,1,\tau}=j,P_{\cdot,1}=\varrho\vert \btheta^{(s)})\right] &= \sum_{i=1}^m\sum_{\cI_i}\bbI(T_{i,1,\tau}=j,P_{i,1}=\varrho)\bbP(\cI_i\vert\cI_i^{obs},\btheta^{(s)}),\nonumber \\
\bbE\left[\bbN(P_{\cdot,t-1}=\varrho_i,P_{\cdot,t}=\varrho_j\vert \btheta^{(s)})\right] &= \sum_{i=1}^m\sum_{\cI_i}\bbI(P_{i,t-1}=\varrho_i,P_{i,t}=\varrho_j)\bbP(\cI_i\vert\cI_i^{obs},\btheta^{(s)}),\nonumber \\
\bbE\left[\bbN(P_{\cdot,t}=\varrho_i,\hat P_{\cdot,t}=\varrho_j\vert \btheta^{(s)})\right] &= \sum_{i=1}^m\sum_{\cI_i}\bbI(P_{i,t}=\varrho_i,\hat P_{i,t}=\varrho_j)\bbP(\cI_i\vert\cI_i^{obs},\btheta^{(s)}),\nonumber \\
\bbE\left[\bbN(T_{\cdot,t-1,\tau}=i,T_{\cdot,t,\tau}=j,P_{\cdot,t}=\varrho\vert \btheta^{(s)})\right] &= \sum_{i=1}^m\sum_{\cI_i}\bbI(T_{i,t-1,\tau}=i,T_{i,t,\tau}=j,P_{i,t}=\varrho)\bbP(\cI_i\vert\cI_i^{obs},\btheta^{(s)}),\nonumber \\
\bbE\left[\bbN(T_{\cdot,t,\tau}=i,\hat T_{\cdot,t,\tau}=j\vert \btheta^{(s)})\right] &= \sum_{i=1}^m\sum_{\cI_i}\bbI(T_{i,t,\tau}=i,\hat T_{i,t,\tau}=j)\bbP(\cI_i\vert\cI_i^{obs},\btheta^{(s)}).\nonumber
\end{align}

To identify the value of $\btheta$ that maximizes the Q-function, we solve the equation by setting the derivatives of the Q-function with respect to $\btheta$ to zero in M-step. 
Considering $\sum_{\varrho\in\cP}\alpha_\varrho=1$, we apply the Lagrange multiplier method to determine the optimal value of $\{\alpha_{\varrho}\}_{\varrho\in\cP}$, resulting in the following equation,
\begin{align}
\label{eqsm:partial1}
\frac{\partial Q(\btheta\vert\btheta^{(s)}) - \lambda(\sum_{\varrho\in\cP}\alpha_\varrho - 1)}{\partial \alpha_\varrho} = 0, (\varrho\in\cP).
\end{align}
Eq.~\eqref{eqsm:partial1} is simplified as follows,
\begin{align}
\label{eqsm:partial2}
\frac{\bbE\left[\bbN(P_{\cdot,1}=\varrho\vert \btheta^{(s)})\right]}{\alpha_\varrho} - \lambda = 0, (\varrho\in\cP).
\end{align}
After taking the sum of Eq.~\eqref{eqsm:partial2} over all $\varrho\in\cP$, we have
\begin{align}
\label{eqsm:partial3}
\sum_{\varrho\in\cP}\bbE\left[\bbN(P_{\cdot,1}=\varrho\vert \btheta^{(s)})\right]=\sum_{\varrho\in\cP} \lambda\alpha_\varrho=\lambda.
\end{align}
When we substitute Eq.~\eqref{eqsm:partial3} into Eq.~\eqref{eqsm:partial2}, we derive the M-step updating equation, Eq.~\eqref{eq:Mstep1}, for the parameters $\balpha=\{\alpha_\varrho\}_{\varrho\in\cP}$.
By applying the similar Lagrange multiplier method to these parameters, we can also derive Eq.~\eqref{eq:Mstep2} through ~\eqref{eq:Mstep6} in the M-step.

This estimation process can be implemented using a standard Baum-Welch algorithm. We employ the standard forward-backward algorithm to compute the six expectations mentioned in Eq.~\eqref{NHMMQfuncDecomp}, given the complexity of enumerating hidden states (refer to Appendix \ref{sec:HMM-imp} for more details).

\subsection{Fast Computation of the E-Step}\label{sec:HMM-imp}
In the parameter estimation steps of HMM-stabilized methods, we use forward-backward algorithm to get the expectation of interest in E-step.
Due to the complexity of hidden state enumeration, we utilize the standard forward-backward algorithm to calculate the expectation in Eq.~\eqref{NHMMQfuncDecomp} via EM algorithm.

For the observation $(\hat{\bP_i},\hat{\bT_i})$ of $n_i$ key frames in video $i$ and parameters $\btheta^{(t)}$ of the HMM-stabilized model, we define the forward and backward variables as follows,
\begin{align}
\bbU_{i,t}(\varrho,\tau)&=\bbP(\hat{\bP}_{i,[n\leq t]},\hat{\bT}_{i,[n\leq t]},P_{i,t}=\varrho,T_{i,t,1}=\tau_1,\cdots,T_{i,t,K}=\tau_{K}),\\
\bbV_{i,t}(\varrho,\tau)&=\bbP(\hat{\bP}_{i,[n> t]},\hat{\bT}_{i,[n>t]}\vert P_{i,t}=\varrho,T_{i,t,1}=\tau_1,\cdots,T_{i,t,K}=\tau_{K})\quad(1\leq t \leq n_i).
\end{align}
where
\[\tau=(\tau_1,\cdots,\tau_{K}),\hat{\bP}_{i,[n\leq t]}=(\hat P_{i,1},\cdots,\hat P_{i,t}),\hat{\bP}_{i,[n>t]}=(\hat P_{i,t+1},\cdots,\hat P_{i,n_i}),\]
\[\hat{\bT}_{i,[n\leq t]}=(\hat T_{i,1},\cdots,\hat T_{i,t}),\hat{\bT}_{i,[n>t]}=(\hat T_{i,t+1},\cdots,\hat T_{i,n_i}).\]

The forward and backward variables can be computed using the following dynamic programming iteration formula,
\begin{align}
\bbU_{i,1}(\varrho,\tau)=&\bbP(P_{i,1}=\varrho)\bbP(\hat P_{i,1}\vert P_{i,1}=\varrho)\prod_{k=1}^{K}\bbP(T_{i,1,k}=\tau_k)\bbP(\hat T_{i,1,k} \vert T_{i,1,k}=\tau_k),\\
\bbU_{i,t+1}(\varrho,\tau)=&\sum_{\varrho^*=1}^{L}\sum_{\tau^*_1=0}^1\cdots\sum_{\tau^*_{K}=0}^1 \bbU_{i,t}(\varrho^*,\tau^*)\bbP(P_{i,t+1}=\varrho\vert P_{i,t}=\varrho^*)\bbP(\hat P_{i,t+1}\vert P_{i,t+1}=\varrho) \nonumber\\
&\prod_{k=1}^{K}\bbP(T_{i,t+1,k}=\tau_k\vert T_{i,t,k}=\tau^*_k)\bbP(\hat T_{i,t+1,k} \vert T_{i,t+1,k}=\tau_k);
\end{align}
\begin{align}
\bbV_{i,n_i}(\varrho,\tau)=&1,\\
\bbV_{i,t-1}(\varrho,\tau)=&\sum_{\varrho^*=1}^{L}\sum_{\tau^*_1=0}^1\cdots\sum_{ \tau^*_{K}=0}^1 \bbV_{i,t}(\varrho^*,\tau^*)\bbP(P_{i,t}=\varrho^*\vert P_{i,t-1}=\varrho)\bbP(\hat P_{i,t-1}\vert P_{i,t-1}=\varrho)\nonumber \\
&\prod_{k=1}^{K}\bbP(T_{i,t-1,k}=\tau^*_k\vert T_{i,t,k}=\tau_k)\bbP(\hat T_{i,t-1,k} \vert T_{i,t-1,k}=\tau_k).
\end{align}
Note that
\begin{align}
\bbP(\hat{\bT},\hat{\bP}\vert\btheta^{(t)})=\sum_{\varrho\in\cP}\sum_{\tau\in\cT}\sum_{i=1}^m\bbU_{i,n_i}(\varrho,\tau).    
\end{align}
The expectations can be computed using the following formulas,
\begin{align}
&\bbE\left[\bbN(P_{\cdot,1}=\varrho)\vert \btheta^{(t)}\right] = \sum_{i=1}^m\sum_{\tau\in\cT}\bbU_{i,1}(\varrho,\tau)\bbV_{i,1}(\varrho,\tau)\bigg/ \bbP(\hat{\bT},\hat{\bP}\vert\btheta^{(t)}),\\
&\bbE\left[\bbN(P_{\cdot,t-1}=\varrho,P_{\cdot,t}=\varrho^*)\vert \btheta^{(t)}\right]= \nonumber\\& \sum_{i=1}^m\sum_{\tau\in\cT}\sum_{\tau^*\in\cT}\sum_{t=1}^{n_i-1}\bbU_{i,t}(\varrho,\tau)\bbV_{i,t+1}(\varrho^*,\tau^*)\bbP(P_{i,t+1}=\varrho^*\vert P_{i,t}=\varrho)\bbP(\hat P_{i,t+1}\vert P_{i,t+1}=\varrho^*) \nonumber \\
&\times\prod_{k=1}^{K}\bbP(T_{i,t,k}=\tau_k\vert T_{i,t+1,k}=\tau^*_k)\bbP(\hat T_{i,t+1,k} \vert T_{i,t+1,k}=\tau^*_k)\bigg/{\bbP(\hat{\bT},\hat{\bP}\vert\btheta^{(t)})},\\ 
&\bbE\left[\bbN(P_{\cdot,t}=\varrho,\hat P_{\cdot,t}=\varrho^*)\vert \btheta^{(t)}\right] = \sum_{\tau\in\cT}\sum_{\tau^*\in\cT}\sum_{(\hat P_{i,t},\hat T_{i,t})=(\varrho^*,\tau^*)}\bbU_{i,t}(\varrho,\tau)\bbV_{i,t}(\varrho,\tau)\bigg/{\bbP(\hat{\bT},\hat{\bP}\vert\btheta^{(t)})},\\
&\bbE\left[\bbN(T_{\cdot,1,k}=\tau)\vert \btheta^{(t)}\right] = \sum_{\varrho\in\cP}\bbU_{i,1}(\varrho,\tau)\bbV_{i,1}(\varrho,\tau)\bigg/ \bbP(\hat{\bT},\hat{\bP}\vert\btheta^{(t)}),\\
&\bbE\left[\bbN(T_{\cdot,t-1,k}=j,T_{\cdot,t,k}=\tau^*,P_{\cdot,t}=\varrho)\vert \btheta^{(t)}\right]= \nonumber\\& \sum_{\varrho^*\in\cP}\sum_{t=1}^{n_i-1}\bbU_{i,t}(\varrho,\tau)\bbV_{i,t+1}(\varrho^*,\tau^*)\bbP(P_{i,t+1}=\varrho^*\vert P_{i,t}=\varrho)\bbP(\hat P_{i,t+1}\vert P_{i,t+1}=\varrho^*)\nonumber \\
&\times\prod_{k=1}^{K}\bbP(T_{i,t,k}=\tau_k\vert T_{i,t+1,k}=\tau^*_k)\bbP(\hat T_{i,t+1,k} \vert T_{i,t+1,k}=\tau^*_k)\bigg/{\bbP(\hat{\bT},\hat{\bP}\vert\btheta^{(t)})},\\
&\bbE\left[\bbN(T_{\cdot,t,k}=\tau,\hat T_{\cdot,t,k}=\tau^*)\vert \btheta^{(t)}\right] = \sum_{\varrho\in\cP}\sum_{\varrho^*\in\cP}\sum_{(\hat P_{i,t},\hat T_{i,t})=(\varrho^*,\tau^*)}\bbU_{i,t}(\varrho,\tau)\bbV_{i,t}(\varrho,\tau)\bigg/{\bbP(\hat{\bT},\hat{\bP}\vert\btheta^{(t)})}.
\end{align}

\subsection{The Degenerated Cases}
When only tool recognition is considered, we get the degenerated likelihood as shown in  Eq.\eqref{eq:HMM-stabilized tools} with parameters $\btheta=(\bbeta,\cA,\cB)$ as the degenerated parameters, resulting in the following Q-function,
\begin{align}
Q(\btheta\vert\btheta^{(s)})&=\sum_{\tau\in\cT}\sum_{j=0}^1((1-j)\log(1-\beta_{\tau})+j\log(\beta_{\tau}))\bbE\left[\bbN(T_{\cdot,1,\tau}=j\vert \btheta^{(s)})\right]\nonumber\\
&+\sum_{\tau\in\cT}\sum_{i=0}^{1}\sum_{j=0}^{1} \log \bA_{\tau}(i,j)\bbE\left[\bbN(T_{\cdot,t-1,\tau}=i,T_{\cdot,t,\tau}=j\vert \btheta^{(s)})\right] \nonumber \\
&+\sum_{\tau\in\cT}\sum_{i=0}^1\sum_{j=0}^1\log \bB_\tau(i,j)\bbE\left[\bbN(T_{\cdot,t,\tau}=i,\hat T_{\cdot,t,\tau}=j\vert \btheta^{(s)})\right],
\end{align}
where $\btheta^{(s)}$ is the parameter estimation in the $s$-th iteration of the EM algorithm, and
\begin{align*}
\bbE\left[\bbN(T_{\cdot,1,\tau}=j\vert \btheta^{(s)})\right] &= \sum_{i=1}^m\sum_{\cI_i}\bbI(T_{i,1,\tau}=j)\bbP(\cI_i\vert\cI_i^{obs},\btheta^{(s)}),\\
\bbE\left[\bbN(T_{\cdot,t-1,\tau}=i,T_{\cdot,t,\tau}=j\vert \btheta^{(s)})\right] &= \sum_{i=1}^m\sum_{\cI_i}\bbI(T_{i,t-1,\tau}=i,T_{i,t,\tau}=j)\bbP(\cI_i\vert\cI_i^{obs},\btheta^{(s)}),\\
\bbE\left[\bbN(T_{\cdot,t,\tau}=i,\hat T_{\cdot,t,\tau}=j\vert \btheta^{(s)})\right] &= \sum_{i=1}^m\sum_{\cI_i}\bbI(T_{i,t,\tau}=i,\hat T_{i,t,\tau}=j)\bbP(\cI_i\vert\cI_i^{obs},\btheta^{(s)}).
\end{align*}

Similarly, when only phase recognition is considered, we have
\begin{align}
Q(\btheta\vert\btheta^{(s)})=&\sum_{\varrho\in\cP}\log\alpha_\varrho\bbE\left[\bbN(P_{\cdot,1}=\varrho\vert \btheta^{(s)})\right]\nonumber\\
&+\sum_{\varrho_i\in\cP}\sum_{\varrho_j\in\cP}\log \bA(\varrho_i,\varrho_j)\bbE\left[\bbN(P_{\cdot,t-1}=\varrho_i,P_{\cdot,t}=\varrho_j\vert \btheta^{(s)})\right]\nonumber \\
&+\sum_{\varrho_i\in\cP}\sum_{\varrho_j\in\cP}\log \bB(\varrho_i,\varrho_j)\bbE\left[\bbN(P_{\cdot,t}=\varrho_i,\hat P_{\cdot,t}=\varrho_j\vert \btheta^{(s)})\right],
\end{align}
where 
\begin{align*}
\bbE\left[\bbN(P_{\cdot,1}=\varrho\vert \btheta^{(s)})\right] &= \sum_{i=1}^m\sum_{\cI_i}\bbI(P_{i,1}=\varrho)\bbP(\cI_i\vert\cI_i^{obs},\btheta^{(s)}),\\
\bbE\left[\bbN(P_{\cdot,t-1}=\varrho_i,P_{\cdot,t}=\varrho_j\vert \btheta^{(s)})\right] &= \sum_{i=1}^m\sum_{\cI_i}\bbI(P_{i,t-1}=\varrho_i,P_{i,t}=\varrho_j)\bbP(\cI_i\vert\cI_i^{obs},\btheta^{(s)}),\\
\bbE\left[\bbN(P_{\cdot,t}=\varrho_i,\hat P_{\cdot,t}=\varrho_j\vert \btheta^{(s)})\right] &= \sum_{i=1}^m\sum_{\cI_i}\bbI(P_{i,t}=\varrho_i,\hat P_{i,t}=\varrho_j)\bbP(\cI_i\vert\cI_i^{obs},\btheta^{(s)}).
\end{align*}

By following the same procedure outlined in Section \ref{sec:HMM-EM}, we can derive the iteration formula of the EM algorithm for the degenerate case. The E-step in the degenerate model can be computed quickly using a similar forward-backward procedure as described in Section \ref{sec:HMM-imp}.

\end{appendices}
\bibliographystylesm{aaai}
\bibliographysm{supplementaryref}
\end{document}